\documentclass[11pt,a4paper]{article}
\usepackage{times,latexsym}
\usepackage{url}
\usepackage[T1]{fontenc}
\usepackage{times}
\usepackage{latexsym}
\usepackage{amsfonts}
\usepackage{amssymb}
\usepackage{amsmath}
\usepackage{booktabs}
\usepackage{graphicx}
\usepackage{caption}
\usepackage{subcaption}

\usepackage{microtype}
\usepackage{fancyhdr,graphicx,amsmath,amssymb}
\usepackage[ruled,vlined]{algorithm2e}
\usepackage{pifont}
\usepackage{multirow}
\newcommand{\xmark}{\ding{55}}
\usepackage{mathabx}

\DeclareMathOperator*{\argmax}{arg\,max}

\usepackage[acceptedWithA]{tacl2018v2}
\usepackage[]{tacl2018v2}

\usepackage{xspace,mfirstuc,tabulary}

\newif\iftaclinstructions
\taclinstructionsfalse 
\iftaclinstructions
\renewcommand{\confidential}{}
\renewcommand{\anonsubtext}{(No author info supplied here, for consistency with
TACL-submission anonymization requirements)}
\newcommand{\instr}
\fi

\iftaclpubformat 

\else

\fi

\DeclareMathOperator*{\topk}{top\_k_2}

\let\emptyset\varnothing

\title{Pretraining the Noisy Channel Model for Task-Oriented Dialogue}

\author{
Qi Liu$^2$\Thanks{Work completed during an internship at DeepMind.}, Lei Yu$^1$, Laura Rimell$^1$, and Phil Blunsom$^{1,2}$ \\
 $^1$DeepMind, $^2$University of Oxford \\
  {\sf \texttt{qi.liu@cs.ox.ac.uk}} \\
  {\sf \texttt{\{leiyu,laurarimell,pblunsom\}@google.com}} \\
}

\date{}

\begin{document}
\maketitle

\begin{abstract}
Direct decoding for task-oriented dialogue is known to suffer from the explaining-away effect, manifested in models that prefer short and generic responses. Here we argue for the use of Bayes' theorem to factorize the dialogue task into two models, the distribution of the context given the response, and the prior for the response itself. This approach, an instantiation of the noisy channel model, both mitigates the explaining-away effect and allows the principled incorporation of large pretrained models for the response prior. We present extensive experiments showing that a noisy channel model decodes better responses compared to direct decoding and that a two stage pretraining strategy, employing both open-domain and task-oriented dialogue data, improves over randomly initialized models.

\end{abstract}

\section{Introduction}

Task-oriented dialogue agents provide a conversational interface to assist users in accomplishing specific goals, such as finding a restaurant or booking a hotel   \cite{seneff2000dialogue,raux2005let,budzianowski2018multiwoz,peng2020soloist}.
Increasing demand from industry for natural language assistants and scalable customer service solutions has recently been driving a renaissance in the development of task-oriented dialogue models.
In addition, the specification of explicit dialogue agent goals, afforded by the task-oriented paradigm, makes such research easier to ground and evaluate than open-domain chatbots.


Current research on task-oriented dialogue is dominated by monolithic sequence-to-sequence models that directly parameterize the conditional distribution of the response given the prior dialogue context. 
However, this monolithic approach conflates the task-specific and language-general aspects of dialogue, and adversely favors short and generic responses \cite{bao2020plato} due to the explaining-away effect \cite{klein2002conditional}.

Here we pursue an alternative to the direct model. Employing Bayes' rule allows us to factorize the probability of the response given the context $p(\mathcal{R} | \mathcal{C})$ into a language model $p(\mathcal{R})$ and a 
context model $p(\mathcal{C} | \mathcal{R})$.\footnote{Here we abstract away from the prediction of belief states and dialogue acts, which also form part of our generative model; see Section \ref{sec-model} for details.}
 Within natural language processing (NLP), this approach is traditionally known as the noisy channel model \cite{shannon48noisy}, and has recently seen renewed interest with its successful application to neural machine translation \cite{yu2016neural,yu2020better,yee2019simple}.

We hypothesize that the noisy channel reformulation is advantageous for dialogue because the factorization enables each sub-module to specialize in a dialogue sub-task. In particular, the context conditional model can help to discount short and generic responses and mitigate the explaining-away effect, while the language model helps ensure that responses are natural. We find that a noisy channel model with the same number of parameters as a direct model achieves better accuracy on three task-oriented dialogue datasets. Moreover, a larger noisy channel model can be trained with the same hardware, by training the sub-modules separately, yielding additional improvements.
It has become common in recent years to pretrain dialogue models on large text data, either general text \cite{peng2020few,budzianowski2019hello,wu2020tod} or dialogue-structured data \cite{roller2020recipes,adiwardana2020towards}, such as tweets and Reddit posts. We utilise a similar strategy with Reddit data and find that the benefits of pretraining to the noisy channel model are similar to those for the direct model. 
Further, we evaluate transfer across task-oriented dialogue datasets by implementing a second pretraining stage using Taskmaster \cite{byrne2019taskmaster} and Schema-Guided Dialogue \cite{rastogi2019towards} as training data, before fine-tuning on our final tasks. 

We evaluate the algorithm on three datasets, MultiWOZ 2.0 \cite{budzianowski2018multiwoz}, CamRest676 \cite{wen2017latent} and SMCalFlow \cite{andreas2020task}, demonstrating that the noisy channel approach is robust to different dialogue schema annotations used across datasets. Further analysis demonstrates that the noisy channel models can decode responses with similar lengths and Zipf scores compared to ground-truth responses and reduce the likelihood of falling into repetition loops \cite{holtzman2019curious}.






\section{A Seq-to-Seq Dialogue Model \label{sec:baseline_model}}

\begin{figure*}[t]
    \centering
    \includegraphics[width=0.86\textwidth, height=193pt]{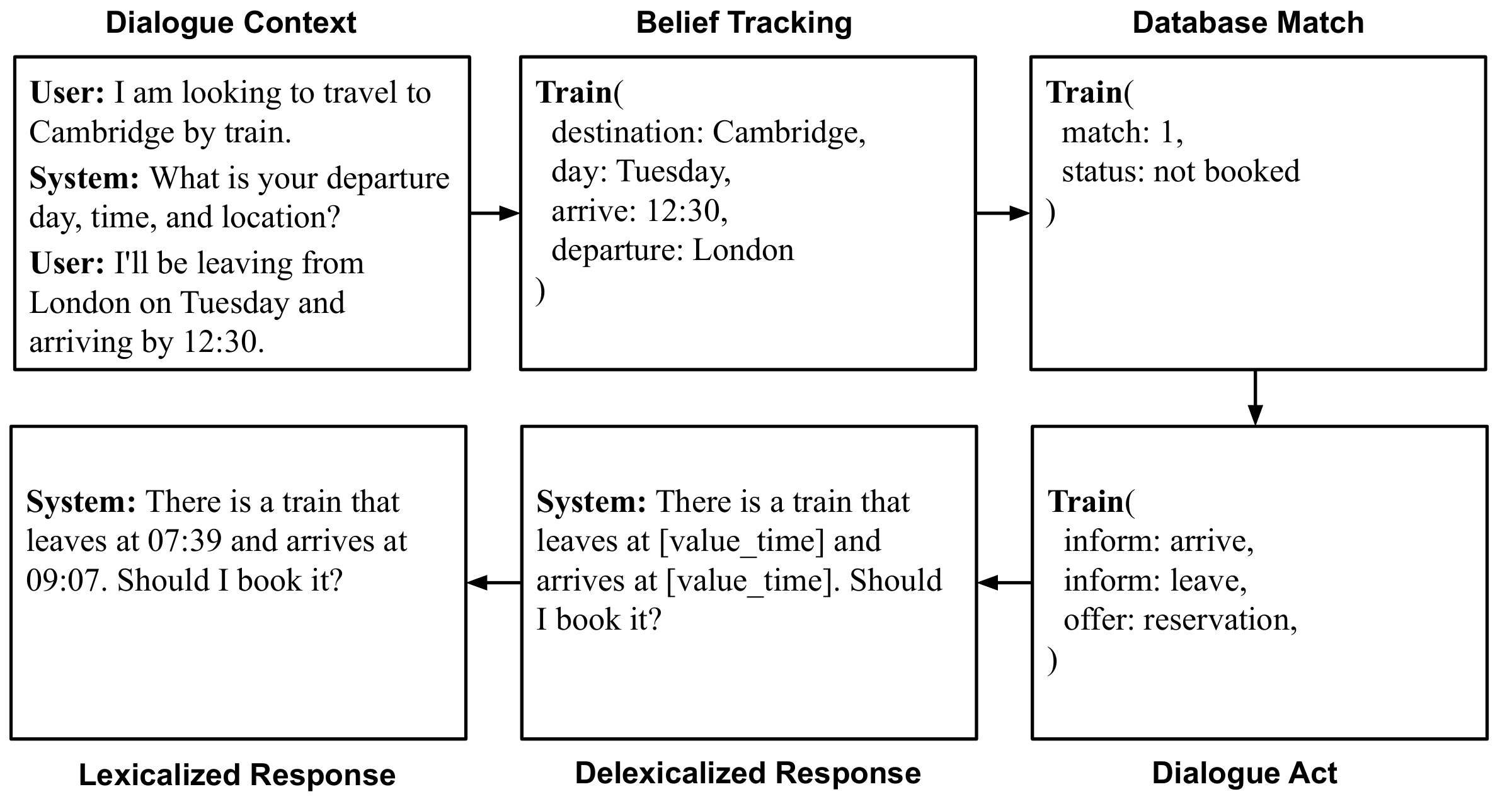}
    \caption{The data flow of one turn in a task-oriented dialogue for train booking from MultiWOZ. }
    \label{fig:dialogue_example}
\end{figure*}

In this section, we introduce a discriminative sequence-to-sequence model for task-oriented dialogue. The traditional sequence of steps needed to produce a system turn in a task-directed dialogue is shown in
Figure \ref{fig:dialogue_example}, with an example from MultiWOZ 2.0 \cite{budzianowski2018multiwoz}. Given a dialogue context containing previous user and system utterances, the dialogue system first predicts a belief state, consisting of a set of slot-value pairs (e.g.\ {\tt destination: Cambridge}), to capture user intent. To ground the system with external information, the belief state can be converted into a database query in order to retrieve relevant information, such as the number of matches and booking information. Next, the system predicts a set of dialogue acts, representing the abstract meaning of the proposed dialogue response \cite{austin1975things}. Finally, a delexicalized dialogue response is generated, where slot values are replaced by generic placeholders, such as {\tt value\_time} for a train departure time, in order to reduce lexical variation. The delexicalized response can be converted to a lexicalized response in post-processing by filling in the slot values based on belief states and database information.

We use the MultiWOZ schema for illustration in Section \ref{sec:baseline_model} and \ref{sec-model}, but our models  easily generalize to different schema annotations (e.g.\ datasets without annotated dialogue acts \cite{andreas2020task}). 

Since it is well known that pipelined models tend to suffer from error propagation, many NLP tasks have been reformulated in recent years as end-to-end text-to-text transformations \cite{raffel2019exploring,brown2020language}. State-of-the-art task-oriented dialogue systems have followed this approach \cite{hosseini2020simple,peng2020few}.
We represent the example from Figure~\ref{fig:dialogue_example} as follows, serializing turns and using special start and end tokens to encapsulate each data field:
\noindent\fbox{%
    \small
    \parbox{\linewidth}{
    Context: \textcolor{black}{[c]} I am looking to ... \textcolor{black}{[/u]} What is your ... \textcolor{black}{[/r]} I'll be leaving ... \textcolor{black}{[/u]} \textcolor{black}{[/c]} \\
    Belief: \textcolor{black}{[b]} [train] destination Cambridge, day Tuesday, arrive 12:30, departure London \textcolor{black}{[/b]} \\
    Database: \textcolor{black}{[db]} [train] match 1, status not booked \textcolor{black}{[/db]} \\
    Act: \textcolor{black}{[a]} [train] inform arrive, inform leave, offer reservation \textcolor{black}{[/a]} \\
    Response: \textcolor{black}{[r]} There is a train that leaves at [value\_time] and arrives at [value\_time]. Should I book it? \textcolor{black}{[/r]}
    }%
}

Given this text representation, the direct discriminative approach models $p(\mathcal{B}, \mathcal{A}, \mathcal{R} | \mathcal{C})$, where $\mathcal{C}$, $\mathcal{B}$, $\mathcal{A}$, and $\mathcal{R}$ represent dialogue context, belief state, dialogue act, and delexicalized response, respectively.\footnote{We do not model the probabilities of database state or lexicalized response, as these are deterministic given the belief state and delexicalized response, respectively.} We use the serialized text of the dialogue context as input, and the concatenation of belief state, dialogue act, and response as target output, making the task amenable to the application of an autoregressive sequence-to-sequence model.
$\mathcal{B}$, $\mathcal{A}$ and $\mathcal{R}$ can be generated sequentially with direct decoding methods, such as greedy decoding and beam search. We use a sequence-to-sequence Transformer \cite{vaswani2017attention} to implement $p(\mathcal{B}, \mathcal{A}, \mathcal{R}| \mathcal{C})$. 
This distribution will also be used to build the noisy channel model in Section \ref{sec-model}.


\section{Noisy Channel Model for Dialogue}
\label{sec-model}

While direct decoding is an effective approach for decoding belief states \cite{hosseini2020simple}, it may be sub-optimal for generating responses. First, it favors short and generic responses \cite{bao2020plato}. As a result, the decoded responses are bland and lack diversity \cite{li2015diversity}. Second, it suffers from the explaining-away effect \cite{klein2002conditional}, where inputs are ``explained-away'' by highly predictive output prefixes. For example, if there is one hotel matching the user's intent as encoded in the belief state, the model is nevertheless prone to decoding ``no'' given the output prefix ``there is'', ignoring the input information. 


In this work, we propose using the neural noisy channel model \cite{yu2016neural} to mitigate the above problems for response generation. Given an input sequence $x$ and output sequence $y$, the noisy channel formulation \cite{shannon48noisy} uses Bayes' rule to rewrite the model $p(y|x)$ as $\frac{p(x|y)p(y)}{p(x)}\ \propto \ p(x|y)p(y)$. It was originally applied to speech recognition, where $p(y|x)$ is a conditional model of the source text given a noisy observation. The {\it channel model} $p(x|y)$ estimates the probability of the observation given the source, while $p(y)$ is an unconditional {\it language model} (or {\it source model}), which can be trained on unpaired data. More recently it has been applied to machine translation, where $y$ is a translation of input text $x$.

Abstracting away from belief states and dialogue acts, for task-oriented dialogue we want to estimate $p(\mathcal{R}|\mathcal{C})$, the probability of a response given a context. The channel model $p(\mathcal{C}|\mathcal{R})$, given a response, predicts a distribution over contexts which might have elicited that response.
The source model $p(\mathcal{R})$ is an unconditional language model.
In this extension of the noisy channel approach to task-oriented dialogue, the ``channel'' can be understood as connecting dialogue contexts with suitable responses. 

For the full task, we develop a noisy channel model for $p(\mathcal{B}, \mathcal{A}, \mathcal{R} | \mathcal{C})$. Using the chain rule, $p(\mathcal{B}, \mathcal{A}, \mathcal{R} | \mathcal{C}) = p(\mathcal{B} | \mathcal{C}) \cdot p(\mathcal{A}, \mathcal{R} | \mathcal{C}, \mathcal{B})$. Following Hosseini-Asl et al.\ \shortcite{hosseini2020simple}, we use the direct model described in  Section~\ref{sec:baseline_model} to parameterize $p(\mathcal{B}|\mathcal{C})$ and decode $\mathcal{B}$, which our preliminary experiments confirmed to be advantageous.

We use the noisy channel formulation to parameterize $p(\mathcal{A}, \mathcal{R} | \mathcal{C}, \mathcal{B})$. Using Bayes' Rule, $p(\mathcal{A}, \mathcal{R} | \mathcal{C}, \mathcal{B})\ \propto\  p(\mathcal{C}, \mathcal{B} | \mathcal{A}, \mathcal{R}) \cdot p(\mathcal{A}, \mathcal{R})$. The channel model $p(\mathcal{C}, \mathcal{B} | \mathcal{A}, \mathcal{R})$ and source model $p(\mathcal{A}, \mathcal{R})$ are implemented as Transformers.




We choose to use the noisy channel formulation for decoding $\mathcal{A}$ based on preliminary experiments which showed improved overall accuracy over direct decoding, possibly because poor dialogue act prediction by the direct model led to worse quality responses. The 
serialized text of $\mathcal{A}$ and $\mathcal{R}$ are concatenated during training, and the decoded sequence is split into $\mathcal{A}$ and $\mathcal{R}$ with the special start/end tokens during decoding.



We suggest that the noisy channel model has three advantages over the direct model for response generation: (1) The channel model can penalize short and generic responses. Such responses can be mapped to a large number of contexts, resulting in a flat distribution over contexts. This leads to a lower channel model score for short and generic responses \cite{zhang2019dialogpt}. (2) The channel model ensures that $(\mathcal{A}, \mathcal{R})$ must explain the corresponding $(\mathcal{C}, \mathcal{B})$, alleviating the explaining-away effect \cite{yu2016neural}. (3) The source model, an unconditional distribution over $\mathcal{A}$ and $\mathcal{R}$, can make use of abundant non-dialogue textual data for pretraining, further improving the fluency of generated sequences \cite{brants2007large}. We leave exploration of this last advantage for future work, as we pretrain all sub-modules with the same data.



\subsection{Decoding}
\label{sec:decoding}

Since exact decoding from the noisy channel model $\argmax_{\mathcal{A}, \mathcal{R}} p(\mathcal{C}, \mathcal{B} | \mathcal{A}, \mathcal{R}) \cdot p(\mathcal{A}, \mathcal{R})$\footnote{Although exact decoding is also computationally intractable for the direct model, approximating $\argmax_{\mathcal{B}} p(\mathcal{B} | \mathcal{C})$ is well-studied, e.g.\ beam search. The decoding for $\mathcal{B}$ is therefore omitted here.} is computationally intractable, we experiment with two approximation methods, noisy channel reranking and noisy channel online decoding. Since these methods rely on $p(\mathcal{A}, \mathcal{R} | \mathcal{C}, \mathcal{B})$ as a proposal distribution for approximation, and both $p(\mathcal{A}, \mathcal{R} | \mathcal{C}, \mathcal{B})$ and $p(\mathcal{B} | \mathcal{C})$ are parameterized with the direct model introduced in Section \ref{sec:baseline_model}, our noisy channel model therefore has three sub-modules: a direct model $p(\mathcal{B}, \mathcal{A}, \mathcal{R} | \mathcal{C})$, a channel model $p(\mathcal{C}, \mathcal{B} | \mathcal{A}, \mathcal{R})$, and a source model $p(\mathcal{A}, \mathcal{R})$.

\begin{algorithm}[t]
\footnotesize
\SetAlgoLined
\SetKwInOut{Input}{Input}
\SetKwInOut{Output}{Output}
\SetKw{Continue}{continue}
\SetKw{Or}{or}
\SetKw{In}{in}
\Input{Context $\mathcal{C}$}
\Output{Belief, act and response $(\mathcal{B}, \mathcal{A}, \mathcal{R})$}
 Decode $\mathcal{B}$ given $\mathcal{C}$ with $p(\mathcal{B}|\mathcal{C})$ \\
 Beam: $\mathcal{S} = \{ (\textcolor{black}{[a]}) \} $ \\
 \While{\normalfont\texttt{end}($\mathcal{S}$) is \textbf{False}}{
  $\mathcal{S'} = \emptyset$ \\
  \For{$\mathcal{O}$ \In $\mathcal{S}$}{
    \If{$\mathcal{O}.\normalfont\texttt{last}$\normalfont()  is \textcolor{black}{[/r]} \Or $\mathcal{|O|}$ $> l$ }{
        $\mathcal{S'}.\normalfont\texttt{add}$($\mathcal{O}$) \\
        \Continue
    }
    Get $k_1$ tokens $o^{1}, ..., o^{k_1}$ from the direct model $p(O_{|\mathcal{O}| + 1} | \mathcal{C}, \mathcal{B}, \mathcal{O})$ \\
    \For{$o^i$ \In $(o^{1}, ..., o^{k_1})$}{
        $\mathcal{S'}.\normalfont\texttt{add}$($(\mathcal{O}, o^i)$)
    }
  }
  $\mathcal{S} = \topk\limits_{\mathcal{O} \in \mathcal{S'}} \log p( \mathcal{O} | \mathcal{C}, \mathcal{B}) + $ \\
  \hspace{32pt} $\lambda_1 \cdot \log p(\mathcal{C}, \mathcal{B}| \mathcal{O} ) + $ \\
  \hspace{54pt} $\lambda_2 \cdot \log p(\mathcal{O}) + $ \\
  \hspace{76pt} $\lambda_3 \cdot |\mathcal{O}| $ \\
 }
 Select $\mathcal{O} \in \mathcal{S}$ with the largest score using Eq.\ \ref{eq:model_combination} and return $(\mathcal{B}, \mathcal{A}, \mathcal{R})$
 \caption{\normalsize Online decoding for the noisy channel. \label{al:noisy_channel}}
\end{algorithm}

\textbf{Noisy channel reranking}: Noisy channel reranking first decodes $\mathcal{B}$ and then continues decoding a list $\mathcal{S}$ of $(\mathcal{A}, \mathcal{R})$ pairs by beam search with the direct model, prior to utilizing the noisy channel model to rerank $(\mathcal{A}, \mathcal{R})$ pairs. In particular, during beam search, partial sequences are expanded and pruned with $p(\mathcal{A}, \mathcal{R} | \mathcal{C}, \mathcal{B})$ (from the direct model in Section \ref{sec:baseline_model}). The pairs after decoding are reranked using the following model combination:
\begin{equation}
\small
\begin{aligned}
\label{eq:model_combination}
(\mathcal{A}', \mathcal{R}') = \argmax\limits_{(\mathcal{A}, \mathcal{R}) \in \mathcal{S}} \log p( \mathcal{A}, \mathcal{R} | \mathcal{C}, \mathcal{B}) & +  \\
\lambda_1 \cdot \log p(\mathcal{C}, \mathcal{B}| \mathcal{A}, \mathcal{R} ) & +  \\
\lambda_2 \cdot \log p(\mathcal{A}, \mathcal{R}) & +  \\
\lambda_3 \cdot |\mathcal{A}, \mathcal{R}| &, \\
\end{aligned}
\end{equation}
where $|\mathcal{A}, \mathcal{R}|$ denotes the length of $(\mathcal{A}, \mathcal{R})$, and $\lambda_1$, $\lambda_2$ and $\lambda_3$ are hyperparameters. Besides the channel model $p(\mathcal{C}, \mathcal{B}| \mathcal{A}, \mathcal{R} )$ and the source model $p(\mathcal{A}, \mathcal{R})$, we additionally use the direct model $p( \mathcal{A}, \mathcal{R} | \mathcal{C}, \mathcal{B})$ and a length bias $|\mathcal{A}, \mathcal{R}|$ to encourage responses with high direct model likelihood and discourage short responses, respectively.

\textbf{Noisy channel online decoding}: In contrast to reranking, online decoding applies the noisy channel model during beam search for pruning partial sequences, thus exploring a larger search space.

As shown in Algorithm \ref{al:noisy_channel}, we first decode the belief state with $p(\mathcal{B} | \mathcal{C})$, which comes from the direct model in Section \ref{sec:baseline_model}. Then, starting with a beam $\mathcal{S}$ containing a single sequence \textcolor{black}{[a]} (the dialogue act start token), we continuously expand the sequences in $\mathcal{S}$ until \texttt{end}($\mathcal{S}$) is met, i.e.\ all sequences in $\mathcal{S}$ either end with \textcolor{black}{[/r]} or have lengths larger than $l$. In each iteration, we first expand the sequences in the beam, then prune the expanded beam. To expand a partial act and response sequence (denoted as $\mathcal{O}$ in Algorithm \ref{al:noisy_channel}), a naive way is to use the noisy channel model to score $|V|$ (the vocabulary size) possible expansions, which is computationally expensive. Instead, we use the probability of the next token $p(O_{|\mathcal{O}| + 1} | \mathcal{C}, \mathcal{B}, \mathcal{O})$ (where $|\mathcal{O}|$ denotes the length of $\mathcal{O}$) to select $k_1$ candidates to be scored by the noisy channel model. This next token probability is from the direct model introduced in Section \ref{sec:baseline_model}. One straightforward way to select $k_1$ expansions from $p(O_{|\mathcal{O}| + 1} | \mathcal{C}, \mathcal{B}, \mathcal{O})$ is using the top-k maximization, but we can also take advantage of the advances in sampling from a categorical distribution for text generation (e.g.\ top-k sampling \cite{fan2018hierarchical} and nucleus sampling \cite{holtzman2019curious}). After the expansion, we prune the expanded beam $\mathcal{S}'$ to obtain a smaller beam with $k_2$ partial sequences based on the model combination in Eq.\ \ref{eq:model_combination}. Compared to noisy channel reranking, online decoding applies the noisy channel model during beam search, which is potentially less biased towards the direct model.

In summary, we note that beam search for both the direct model and the online decoding for our noisy channel model decodes ($\mathcal{B}, \mathcal{A}, \mathcal{R}$) autoregressively. Thus both approaches are end-to-end models for task-oriented dialogue. The key difference is that noisy channel online decoding uses Eq.\ \ref{eq:model_combination} for pruning, while the direct model uses $p(\mathcal{A}, \mathcal{R} | \mathcal{C}, \mathcal{B})$.

\section{Model and Pretraining}

We use three Transformer \cite{vaswani2017attention} networks to parameterize the direct model $p(\mathcal{B}, \mathcal{A}, \mathcal{R} | \mathcal{C})$, the channel model $p(\mathcal{C}, \mathcal{B} | \mathcal{A}, \mathcal{R})$ and the source model $p(\mathcal{A}, \mathcal{R})$, respectively. The input to each Transformer is the sum of four embeddings: word embeddings, position embeddings, role embeddings (user/system), and turn embeddings (each word corresponds to a turn number). Cross entropy is used as the loss function.

Given training samples $(\mathcal{C}, \mathcal{B}, \mathcal{A}, \mathcal{R})$, if we train the channel model using complete $(\mathcal{A}, \mathcal{R})$ pairs as input, a significant discrepancy arises between training and decoding for noisy channel online decoding. Since the channel model is used to score partial act and response pairs, i.e.\ $p(\mathcal{C}, \mathcal{B}| \mathcal{O})$ in Algorithm \ref{al:noisy_channel}, the channel model trained with complete $(\mathcal{A}, \mathcal{R})$ pairs is unsuited to scoring partial sequences.
In order to manually create partial sequences during training that are better matched for online decoding, we truncate the $(\mathcal{A}, \mathcal{R})$ pairs with a truncation length uniformly sampled from 1 to the sequence length (inclusive). The direct model and the source model are trained with complete sequences, as partial sequences occur naturally in their standard autoregressive training  procedure.


As in-domain dialogue data are usually scarce, we use a two-stage pretraining strategy to enhance the noisy channel model. Although the effectiveness of pretraining with Reddit data has been validated for open-domain dialogue \cite{zhang2019dialogpt,bao2019plato,adiwardana2020towards}, relatively little work has applied such data to task-oriented dialogue.\footnote{One exception is Henderson et al.\ \shortcite{henderson2019training}, who use Reddit data to improve response retrieval and selection. We focus on response generation in this work.} In the first stage, we explore Reddit pretraining (where the Reddit data is pre-processed into $(\mathcal{C}, \mathcal{R})$, i.e.\ context-response, pairs as described below). In the second stage, we use two task-oriented dialogue datasets, Taskmaster\footnote{\url{https://cutt.ly/xkuUHUa}} \cite{byrne2019taskmaster} and Schema-Guided Dialogue\footnote{\url{https://cutt.ly/QkuUZUu}} \cite{rastogi2019towards}, to specialize the Reddit-pretrained models. Since the Reddit data consists of open-domain-style dialogues (where belief states and dialogue acts are missing), pretraining on these datasets can familiarize the models with the sequence-to-sequence representation of task-oriented dialogue. 
Three models, a context-to-response model, a response-to-context model and a response language model, are pretrained to initialize the direct model, the channel model and the source model, respectively.

\begin{table*}[t]
    \centering
    \footnotesize
    \renewcommand{\arraystretch}{0.9}
    \setlength{\tabcolsep}{3pt}
    \scalebox{0.96}{
    \begin{tabular}{lrrrrrcrr}
        \toprule
            \textbf{Dataset} & \textbf{\# Dialog} & \textbf{\# Turn} & \textbf{Avg. Turn/Dialog} & \textbf{Avg. Token/Turn} & \textbf{\# Domain} & \textbf{Multi-Task} & \textbf{\# Unique Slot} & \textbf{\# Unique Value} \\ \midrule
            Taskmaster & 17,304 & 341,801 & 19.75 & 7.87 & 7 & \xmark & 281 & 66,659\\
            Schema & 22,825 & 463,284 & 20.3 & 9.86 & 17 & \checkmark &  123 & 23,889\\
            CamRest676 & 676 & 5,488 & 8.12 & 10.71 & 1 & \xmark & 4 & 89\\
            MultiWOZ & 10,438 & 143,048 & 13.7 & 15.03 & 7 & \checkmark & 46 & 11,828\\
            SMCalFlow & 41,517 & 170,590 & 4.11 & 8.77 & 4 & \checkmark & - & - \\
        \bottomrule
    \end{tabular}
    }
    \caption{Statistics of task-oriented dialogue datasets. We define a multi-task dialogue as a dialogue involving multiple tasks, e.g.\ hotel and restaurant booking, while its counterpart handles a single task, e.g.\ hotel booking. Taskmaster and CamRest676 do not contain any multi-task dialogues. \label{tab:dataset_stats}}
\end{table*}

\subsection{Implementation Details}

\textbf{Models}: All models are implemented with JAX \cite{jax2018github} and Haiku \cite{haiku2020github}. For the direct model introduced in Section \ref{sec:baseline_model}, we use a Transformer model with hidden size 512, 12 encoder-decoder layers, and 16 self-attention heads. The model has 114M parameters. For the noisy channel model, we use a base setting and a large setting. The base setting reduces the number of layers to 5, hidden size to 384 and self-attention heads to 12. Its sub-modules, a direct model, a reverse model and a language model, have 43M, 43M and 30M parameters, respectively. We employ the base setting for a fair comparison with a single direct model using roughly the same number of parameters (116M vs. 114M). For the large setting, we use the same hyperparameters as the direct model (114M), so that its sub-modules, a direct model, a reverse model and a language model, have 114M, 114M and 64M parameters, respectively. We use this large setting to explore the limits of the noisy channel model. The large noisy channel model (292M) is 2.56 times larger compared to the direct model (114M). This illustrates another advantage of the noisy channel model during training. While training a direct model with 292M parameters will overflow the memory of 16GB TPUs (v3) without using model parallelism, training the sub-modules of the large noisy channel model can easily fit into 16GB TPUs, as these modules are independently trained with no need to load three modules for training. This enables us to train a noisy channel model with more parameters compared to training a direct model using the same hardware. For inference, we still need to load the sub-modules into a TPU. Since gradients are not required during inference, we are able to load the three sub-modules of the large noisy channel model (292M) into a single TPU with 16GB memory for decoding. The large noisy channel model (292M) still consumes more memory than the direct model (114M) during inference.

\textbf{Pretraining settings}: The maximum sequence length $l$ is set to 1024, and sequences with longer lengths are truncated. We reuse the vocabulary from GPT-2 \cite{radford2019language}, which contains 50,257 BPE tokens. We use PreNorm \cite{nguyen2019transformers} for faster convergence. GELU \cite{hendrycks2016gaussian} is applied as the activation function. Following ALBERT \cite{lan2019albert}, dropout is disabled during pretraining. We use the normal distribution truncated to the range $[-0.01, 0.01]$ to initialize the input embeddings, while other parameters are initialized using the normal distribution with zero mean and standard deviation 0.1. The batch size is set to 256. The LAMB optimizer \cite{you2019large} ($b_1=0.9$ and $b_2=0.999$) is employed for optimization. The initial learning rate is 1e-7, and we apply 4000 warmup steps to increase the learning rate to 1e-3, before utilizing cosine annealing to decay the learning rate. Gradient clipping with clipping value 1 is applied to avoid gradient explosion. We use gradient accumulation with accumulation step 20. 

\textbf{Pretraining}: For Reddit pretraining, we download a Reddit dump (with Reddit posts ranging from 2005-12 to 2019-09) from PushShift.\footnote{\url{https://pushshift.io/}} Since the comments of a Reddit post are organized into a tree, we extract paths from a tree as dialogue turns. The last comment of each comment path is regarded as the response, while the others are used as the dialogue context. We pretrain each model for 400,000 steps, consuming 102,400,000 (400,000 $\times$ 256) comment paths in total. For the task-oriented pretraining, we combine the two datasets, Taskmaster and Schema-Guided Dialogue, and pretrain for 1e5 steps. The statistics of the task-oriented dialogue datasets are shown in Table \ref{tab:dataset_stats}.

We train each model using 64 TPU chips with 16GB memory each. The pretraining takes around 4 days to complete. 

\section{Experiments}

We fine-tune and evaluate the pretrained models on three dialogue datasets: MultiWOZ 2.0, CamRest676 and SMCalFlow \cite{andreas2020task}. In this section we describe the datasets (Section \ref{sec:datasets}), fine-tuning (Section \ref{sec:fine_tuning}), decoding (Section \ref{sec:decoding_method}) and evaluation metrics (Section \ref{sec:evaluation_metrics}). Results are presented in Section~\ref{sec:results}, and analysis and ablation studies in Section~\ref{sec:analysis}. 

\begin{table*}[t]
    \centering
    \footnotesize
    \renewcommand{\arraystretch}{0.9}
    \begin{tabular}{lcccc}
        \toprule
        \textbf{Model}   & \textbf{Inform $\uparrow$} & \textbf{Success $\uparrow$} &
        \textbf{BLEU $\uparrow$} &
        \textbf{Combined $\uparrow$} \\ \midrule
        Sequicity \cite{lei2018sequicity} & 66.4 & 45.3 & 15.54 & 71.39\\
        HRED-TS \cite{peng2019teacher} & 70.0 & 58.0 & 17.50 & 81.50 \\
        DSTC8 Track 1 Winner \cite{ham2020end} & 73.0 & 62.4 & 16.00 & 83.50 \\
        DAMD \cite{zhang2019task} & 76.4 & 60.4 & 16.60 & 85.00 \\
        SimpleTOD \cite{hosseini2020simple} & 84.4 & 70.1 & 15.01 & 92.26 \\ 
        SOLOIST \cite{peng2020soloist} & 85.5 & 72.9 & 16.54 & 95.74 \\
        UBAR \cite{yang2020ubar}$^\dagger$ & \textbf{88.2} & \textbf{79.5} & 16.43 & 100.28 \\
        \midrule\multicolumn{5}{c}{Randomly Initialized} \\ \midrule
        Direct decoding (114M) & 81.0 & 54.7 & 15.12 & 82.97 \\
        Noisy channel reranking (116M) & 82.7 & 57.1 & 15.29 & 85.19\\
        Noisy channel online decoding (116M) & 82.9 & 58.9 & 15.33 & 86.23\\
        Noisy channel reranking (292M) & 82.1 & 58.1 & 15.37 & 85.47 \\
        Noisy channel online decoding (292M) & \textbf{83.9} & \textbf{60.9} & \textbf{15.57} & \textbf{87.97} \\
        \midrule\multicolumn{5}{c}{Reddit Pretraining} \\ \midrule
        Direct decoding (114M) & 81.0 & 69.2 & 17.06 & 92.16 \\
        Noisy channel reranking (116M) & 81.3 & 70.1 & 19.01 & 94.71\\
        Noisy channel online decoding (116M) & 81.6 & 71.1 & 19.31 & 95.66\\
        Noisy channel reranking (292M) & 82.2 & 70.9 & 19.89 & 96.44 \\
        Noisy channel online decoding (292M) & \textbf{82.4} & \textbf{71.7} & \textbf{20.49} & \textbf{97.54} \\
 \midrule
        \multicolumn{5}{c}{Task-Oriented Pretraining} \\ \midrule
        Direct decoding (114M) & 85.2 & 72.9 & 17.00 & 96.05 \\
        Noisy channel reranking (116M) & 85.6 & 73.8 & 19.38 & 99.08\\
        Noisy channel online decoding (116M) & 85.9 & 74.8 & 19.76 & 100.11\\
        Noisy channel reranking (292M) & 86.5 & 74.9 & 20.31 & 101.01 \\
        Noisy channel online decoding (292M) & \textbf{86.9} & \textbf{76.2} & \textbf{20.58} & \textbf{102.13} \\
    \bottomrule
    \end{tabular}
    \caption{MultiWOZ test results (end-to-end modeling with generated beliefs) with seq2seq approaches. Results are significant (p < 0.01) comparing noisy channel decoding and direct decoding. $\dagger$ \citet{yang2020ubar} also report a combined score of 105.1 with an alternative context and evaluation setting, contributions orthogonal to our work and the other benchmarks reported here.} 
    \label{tab:multiwoz_end2end}
\end{table*}

\subsection{Datasets \label{sec:datasets}}


MultiWOZ\footnote{\url{https://cutt.ly/0kuUCRS}} is a multi-domain dataset consisting of dialogues annotated with $\mathcal{C}, \mathcal{B}, \mathcal{A}, \mathcal{R}$ in the following seven domains: attraction, hotel, hospital, police, restaurant, train, and taxi. Since its release, MultiWOZ has been one of the most commonly used task-oriented dialogue datasets. 

CamRest676\footnote{\url{https://cutt.ly/SkuUNfE}} is annotated similarly to MultiWOZ and consists of dialogues in a single domain: restaurant reservations. Though CamRest676 is smaller than MultiWOZ and predates it, it still provides a widely used benchmark for evaluating task-oriented dialogue models. 

SMCalFlow consists of dialogues in four domains: calendar, weather, places, and people. Unlike MultiWOZ and CamRest676, SMCalFlow uses dataflow graphs instead of slot-value pairs to represent belief states and does not annotate dialogue acts. We refer readers to Andreas et al.\ \shortcite{andreas2020task} for a detailed description of the dataflow representation. We follow Andreas et al.\ \shortcite{andreas2020task} to convert dataflow graphs into sequences to apply seq2seq models. This dataset is newer and offers fewer prior models to compare with, but we use this dataset to study the robustness of the noisy channel model under different annotation schemas. 

We use the public splits for these datasets, where MultiWOZ, CamRest676 and SMCalFlow are split to 8438/1000/1000, 404/136/136 and 32647/3649/5211 dialogues for training, development and testing, respectively. However, since SMCalFlow's test set has not been publicly released, we randomly select 500 dialogues from its training set to tune hyperparameters and use its development set for testing.

\textbf{Preprocessing}: We use the standard preprocessing procedures for each dataset in order to facilitate fair comparison with previous methods.\footnote{\url{https://cutt.ly/TkuU1oM}} \footnote{\url{https://cutt.ly/zkuU0Ht}} \footnote{\url{https://cutt.ly/vkuU9bT}} In particular, for MultiWOZ and CamRest676, delexicalization is used to reduce lexical variation, while SMCalFlow does not use delexicalization. During delexicalization, slot values are replaced by generic placeholders based on a pre-defined dictionary. 
During decoding, following prior work, our dialogue models generate delexicalized responses. These delexicalized responses are re-lexicalized in post-processing by replacing placeholders with their corresponding slot values based on belief states and database information. Since there is no public code for lexicalization,\footnote{We confirmed this with the dataset authors by email.} we implement our own functions for lexicalization with regular expressions, for the purpose of displaying example responses. However, this does not affect reported results, as the standard metrics for MultiWOZ and CamRest676 which we adopt here are calculated using delexicalized responses.

\begin{table*}[t]
    \centering
    \footnotesize
    \renewcommand{\arraystretch}{0.9}
    \begin{tabular}{lcccc}
        \toprule
        \textbf{Model}   & \textbf{Inform $\uparrow$} & \textbf{Success $\uparrow$} &
        \textbf{BLEU $\uparrow$} &
        \textbf{Combined $\uparrow$} \\ \midrule
        Sequicity \cite{lei2018sequicity} & 92.3 & 85.3 & 21.40 & 110.20 \\
        GPT-2 fine-tuned \cite{wu2019alternating} & - & 86.2 & 19.20 & - \\
        ARDM \cite{wu2019alternating} & - & \textbf{87.1} & 25.20 & - \\
        SOLOIST \cite{peng2020soloist} & 94.7 & \textbf{87.1} & 25.50 & 116.40 \\ \midrule
       \multicolumn{5}{c}{Randomly Initialized} \\ \midrule
        Direct decoding (114M) & 78.1 & 83.5 & 21.58 & 102.38 \\
        Noisy channel online decoding (116M) & 79.8 & 84.1 & 22.83 & 104.78\\
        Noisy channel online decoding (292M) & \textbf{80.9} & \textbf{84.9} & \textbf{23.19} & \textbf{106.09} \\
        \midrule\multicolumn{5}{c}{Reddit Pretraining} \\ \midrule
        Direct decoding (114M) & 93.3 & 83.9 & 23.41 & 112.01 \\
        Noisy channel online decoding (116M) & 93.7 & 84.5 & 25.14 & 114.24\\
        Noisy channel online decoding (292M) & \textbf{93.9} & \textbf{84.7} & \textbf{25.38} & \textbf{114.68} \\
 \midrule
        \multicolumn{5}{c}{Task-Oriented Pretraining} \\ \midrule
        Direct decoding (114M) & 93.4 & 84.3 & 24.92 & 113.77 \\
        Noisy channel online decoding (116M) & 94.3 & 85.2 & 25.98 &	115.73\\
        Noisy channel online decoding (292M) & \textbf{95.4} & \textbf{85.3} & \textbf{26.89} & \textbf{117.24} \\
    \bottomrule
    \end{tabular}
    \caption{CamRest676 test results (end-to-end modeling with generated beliefs) with seq2seq approaches. Noisy channel reranking performs comparable with noisy channel online decoding, and the results are not shown. Results are significant (p < 0.01) comparing noisy channel decoding and direct decoding. \label{tab:camrest676_end2end}}
\end{table*}
\begin{table}[t]
    \centering
    \footnotesize
    \renewcommand{\arraystretch}{0.9}
    \begin{tabular}{lcc}
        \toprule
        \textbf{Model}   & \textbf{SacreBLEU $\uparrow$} & \textbf{TER $\downarrow$}  \\ \midrule
       \multicolumn{3}{c}{Randomly Initialized} \\ \midrule
        Direct decoding (114M) & 51.30 & 89.13\\
        Online decoding (116M) & 53.66 & 74.18\\
        Online decoding (292M) & \textbf{54.39} & \textbf{73.18}\\
        \midrule\multicolumn{3}{c}{Reddit Pretraining} \\ \midrule
        Direct decoding (114M) & 60.68 & 61.99\\
        Online decoding (116M) & 63.29 & 47.16\\
        Online decoding (292M) & \textbf{63.91} & \textbf{46.43}\\
 \midrule
        \multicolumn{3}{c}{Task-Oriented Pretraining} \\ \midrule
        Direct decoding (114M) & 61.02 & 59.84\\
        Online decoding (116M) & 63.72 & 46.27\\
        Online decoding (292M) & \textbf{64.29} & \textbf{45.81}\\
    \bottomrule
    \end{tabular}
    \caption{SMCalFlow results. Reranking performs worse than online decoding, and the results are not shown. Results are significant (p < 0.01) comparing noisy channel decoding and direct decoding. \label{tab:semantic_machine_end2end}}
\end{table}

\subsection{Fine-Tuning \label{sec:fine_tuning}}

We apply label smoothing with parameter 0.1. Dropout is used on input embeddings and hidden representations, with dropout rate 0.1. The Adam optimizer \cite{kingma2014adam} ($b_1=0.9$ and $b_2=0.999$) is adopted. We use a fixed learning rate 1e-4 with gradient clipping for fine-tuning. 

\subsection{Decoding \label{sec:decoding_method}}
We use direct decoding for belief state. For dialogue act and response, we study three decoding methods: direct decoding, noisy channel reranking and noisy channel online decoding. Since all of these decoding methods require choosing $k_1$ tokens from a categorical distribution during expansion, we compare four methods, top-k maximization, sampling without replacement, top-k sampling, and nucleus sampling. Nucleus sampling with cumulative probability 0.98 performs marginally better and is adopted. We perform a range search with the range $[1, 20]$ on development sets for the beam sizes $k_1$ and $k_2$, and we set $k_1, k_2 = 4$, $k_1, k_2 = 15$ and $k_1, k_2 = 4$ for MultiWOZ, CamRest676 and SMCalFlow, respectively. For noisy channel reranking and noisy channel online decoding, a grid search with range $[0, 2]$ is performed for $\lambda_1$, $\lambda_2$ and $\lambda_3$. We set ($\lambda_1=0.8$, $\lambda_2=1$, $\lambda_3=0.8$),  ($\lambda_1=1.2$, $\lambda_2=1.2$, $\lambda_3=0.8$) and ($\lambda_1=0.4$, $\lambda_2=1$, $\lambda_3=0.2$) for MultiWOZ, CamRest676 and SMCalFlow, respectively.

\subsection{Evaluation Metrics \label{sec:evaluation_metrics}}

For MultiWOZ and CamRest676, following previous work, we adopt three automatic evaluation metrics: inform, success and BLEU score. 
\citet{peng2020soloist} showed that these metrics are well correlated to human evaluation.
The evaluators\footnote{\url{https://cutt.ly/VkuU3FA}} \footnote{\url{https://cutt.ly/MkuU88u}} provided with the datasets are used for calculating these metrics. To calculate the inform score for a dialogue, the evaluator first checks whether certain placeholders (e.g.\ {\tt [restaurant\_name])} appear in decoded responses. If so, decoded belief states are converted to database queries to retrieve database records. These database records are compared with the records retrieved with ground-truth belief states. The inform score is one if these two sets of database records match. The success score takes all the requestable slots (e.g.\ postcode, phone number and address) from a decoded response and compares these requestable slots with the ones in the ground-truth response. The success score is one if generated requestable slots coincide with the ground-truth ones. BLEU score (BLEU-4) compares the n-grams of generated responses and human responses, and is a widely used metric in NLP for evaluating text quality. Following \citet{budzianowski2018multiwoz}, we also calculate a combined score, which is (Inform + Success) / 2 + BLEU. For SMCalFlow, inform and success scores are not applicable since calculation of these scores relies on delexicalization placeholders, and this dataset does not use delexicalization.
We use SacreBLEU\footnote{\url{https://cutt.ly/BkuU7dL}} and TER\footnote{\url{https://pypi.org/project/pyter/}} to directly measure the quality of responses. As prior work on this dataset has focused on belief tracking rather than end-to-end response generation, we are the first to use these metrics on this dataset.

We perform significance tests, where we use t-test for inform, success and TER scores and use permutation test for BLEU. 


\section{Results}
\label{sec:results}





{\bf MultiWOZ:}
Results on the MultiWOZ test set are shown in Table \ref{tab:multiwoz_end2end}. We observe several trends. First, the base noisy channel model (116M) performs  better than direct decoding (114M), despite having a similar number of parameters, showing that the noisy channel factorization is beneficial for task-oriented dialogue. The large noisy channel setting improves further over the base setting. Second, Reddit pretraining provides benefits over random initialization, validating the use of large open-domain dialogue-genre pretraining for task-oriented dialogue, while the models with a second stage of task-oriented pretraining obtain further improvements. This effect is consistent across both direct and noisy channel decoding. Finally, we observe that online decoding consistently outperforms reranking, indicating the benefits of tighter model integration during decoding.


Our model performs better on combined score than SOLOIST \cite{peng2020soloist}, a closely related baseline which pretrains a GPT2-initialized Transformer with Taskmaster and Schema-Guided Dialogue and decodes with nucleus sampling.

{\bf CamRest676:}
Results on the CamRest676 test set are shown in Table \ref{tab:camrest676_end2end}. We observe that the base noisy channel model (116M) obtains better results compared to direct decoding (114M), again demonstrating the effectiveness of the noisy channel model. Reddit pretraining again provides a large benefit over random initialization for both direct decoding and noisy channel decoding, while task-oriented pretraining provides a further boost. Our model again performs better than SOLOIST.






{\bf SMCalFlow:}
Results on the SMCalFlow development set are shown in Table~\ref{tab:semantic_machine_end2end}. As end-to-end models have not previously been tested on this dataset, we use it to demonstrate that the noisy channel model, which we developed primarily on MultiWOZ, continues to be effective on task-oriented dialogue datasets with different annotation schema. The results are consistent with MultiWOZ and CamRest676. The noisy channel model outperforms the direct model by a large margin, demonstrating that dialogue act annotations are not essential for the noisy channel model, and that it remains effective across diverse dialogue representations.

Reddit pretraining confers a similar large benefit on SMCalFlow as on the other datasets, but we observe that task-oriented pretraining brings only marginal further improvements. This may be due to differences in domain or format between our pretraining datasets and SMCalFlow. Alternatively, task-oriented pretraining may help more on task-specific metrics, such as inform and success scores, than on text quality metrics such as BLEU and TER scores. This hypothesis is further supported by the MultiWOZ results in Table~\ref{tab:multiwoz_end2end}.


\section{Analysis}
\label{sec:analysis}

In this section, we use MultiWOZ and CamRest676 to perform ablation studies on the effects of model combination, large-scale pretraining, and sample efficiency; as well as analyzing the runtime requirements of our model and the reasons for its success.


\begin{table}[t]
    \centering
    \footnotesize
    \renewcommand{\arraystretch}{0.9}
    \begin{tabular}{lcc}
        \toprule
        \textbf{Model}   & \textbf{CamRest676} & \textbf{MultiWOZ} \\ \midrule
    Direct decoding & 115.17 & 96.73   \\ \midrule
    \multicolumn{3}{c}{Noisy Channel Online Decoding} \\ \midrule
    Direct + Channel & 115.63 & 98.54\\
    Direct + Source & 115.91 & 99.12\\
    Direct + Length & 115.56 & 97.57\\
    Channel + Source & 115.82 & 99.18\\ 
    Channel + Length & 115.60 & 98.71\\
    Source + Length & 115.62 & 99.19\\
    All - Direct & 115.96 & 100.89\\
    All - Channel & 116.56 & 100.93\\
    All - Source & 116.38 & 99.92\\
    All - Length & 116.52 & 101.11\\
    All & \textbf{116.91} & \textbf{102.62}\\
    \bottomrule
    \end{tabular}
    \caption{Ablation results for model combination on development sets (combined score). Results for reranking are similar and are not shown. `All', `Direct' `Source', and `Channel' denote no ablation, direct model, source model and channel model, respectively. Rows with `+' are combinations of two sub-modules, while the rows with `-' are combinations of three sub-modules. \label{tab:ablation_study}}
\end{table}

\begin{figure*}[t]
    \centering 
    \hspace{-2em}
    \begin{subfigure}[b]{0.5\textwidth}
        \centering 
        \includegraphics[height=82pt]{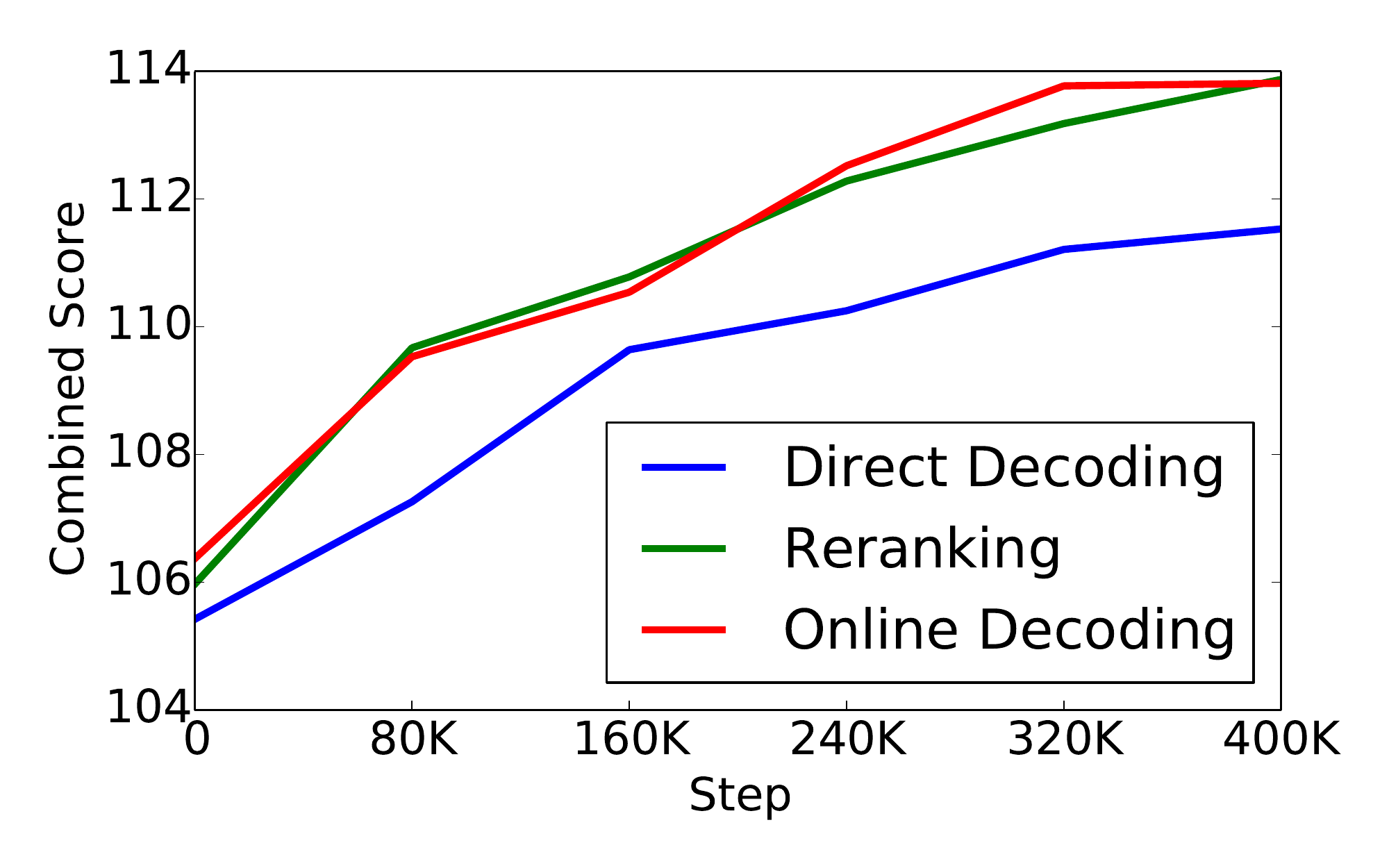}
        \vspace{-1.0em}
        \caption{Reddit pretraining, CamRest676}    
    \end{subfigure} \hspace{-2em}
    \begin{subfigure}[b]{0.5\textwidth}
        \centering 
        \includegraphics[height=82pt]{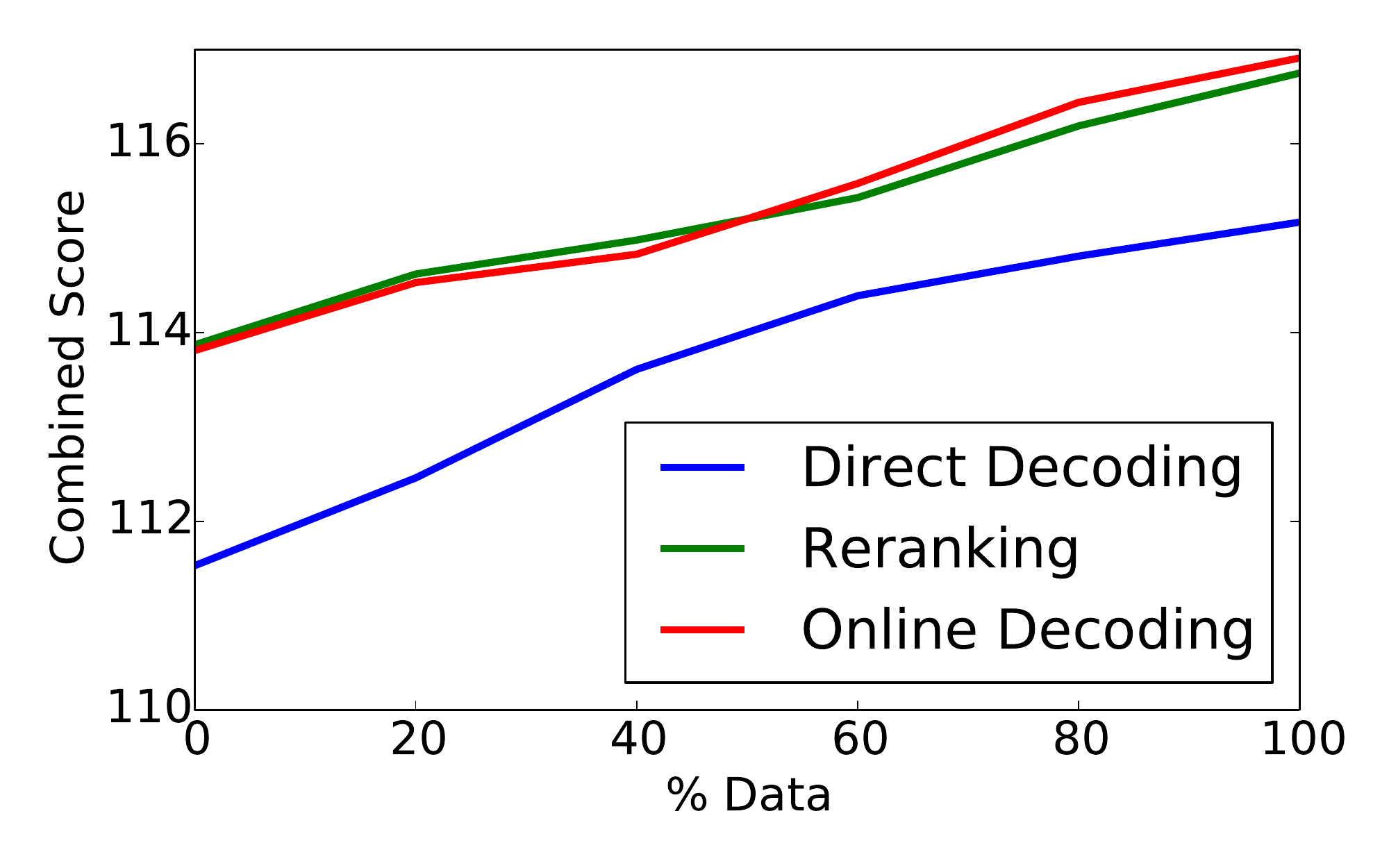}
        \vspace{-1.0em}
        \caption{Task-oriented pretraining, CamRest676}  
    \end{subfigure} \hspace{-2em}

    \begin{subfigure}[b]{0.5\textwidth}
        \centering 
        \includegraphics[height=82pt]{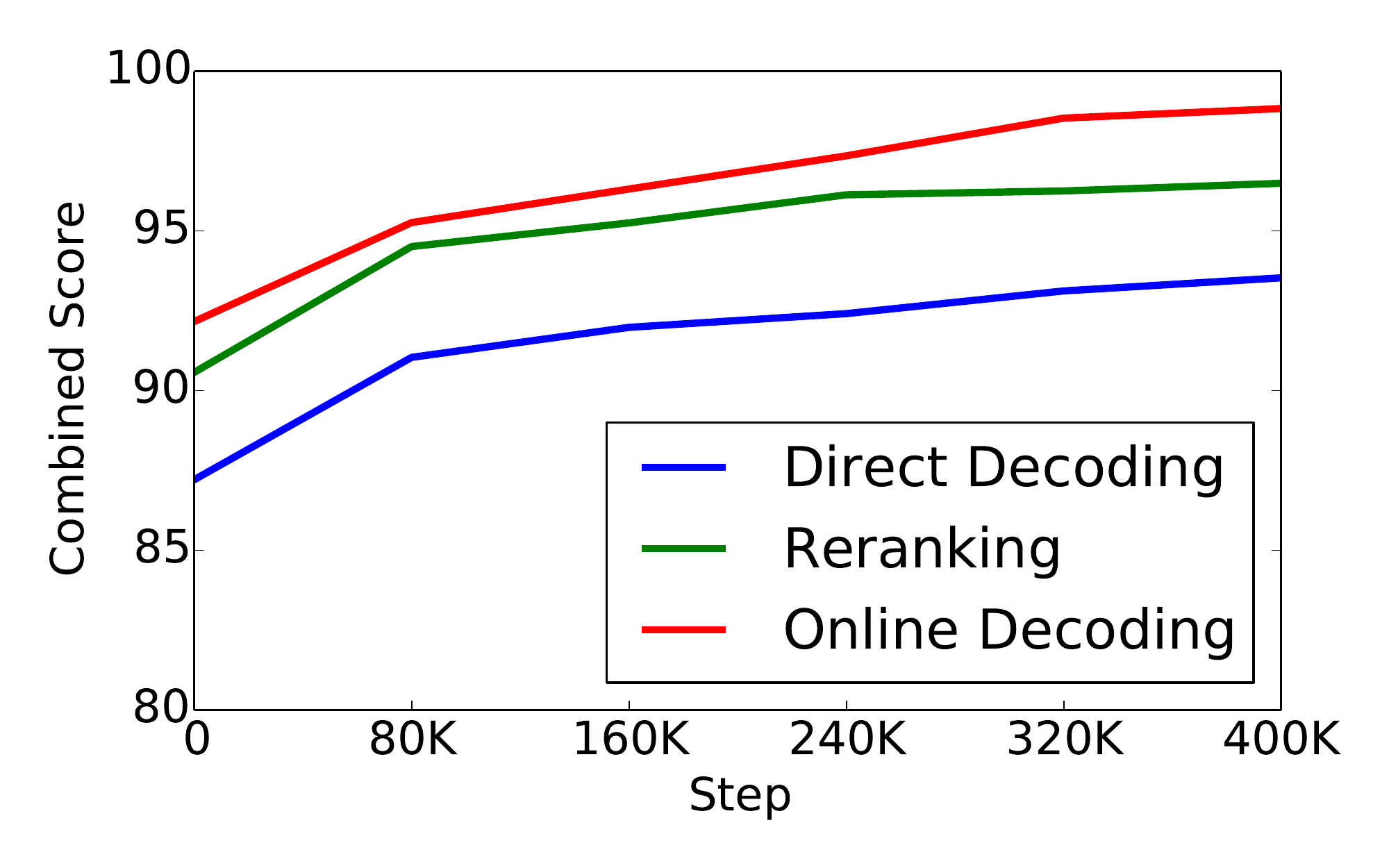}
        \vspace{-1.0em}
        \caption{Reddit pretraining, MultiWOZ}    
    \end{subfigure} \hspace{-2em}
    \begin{subfigure}[b]{0.5\textwidth}
        \centering 
        \includegraphics[height=82pt]{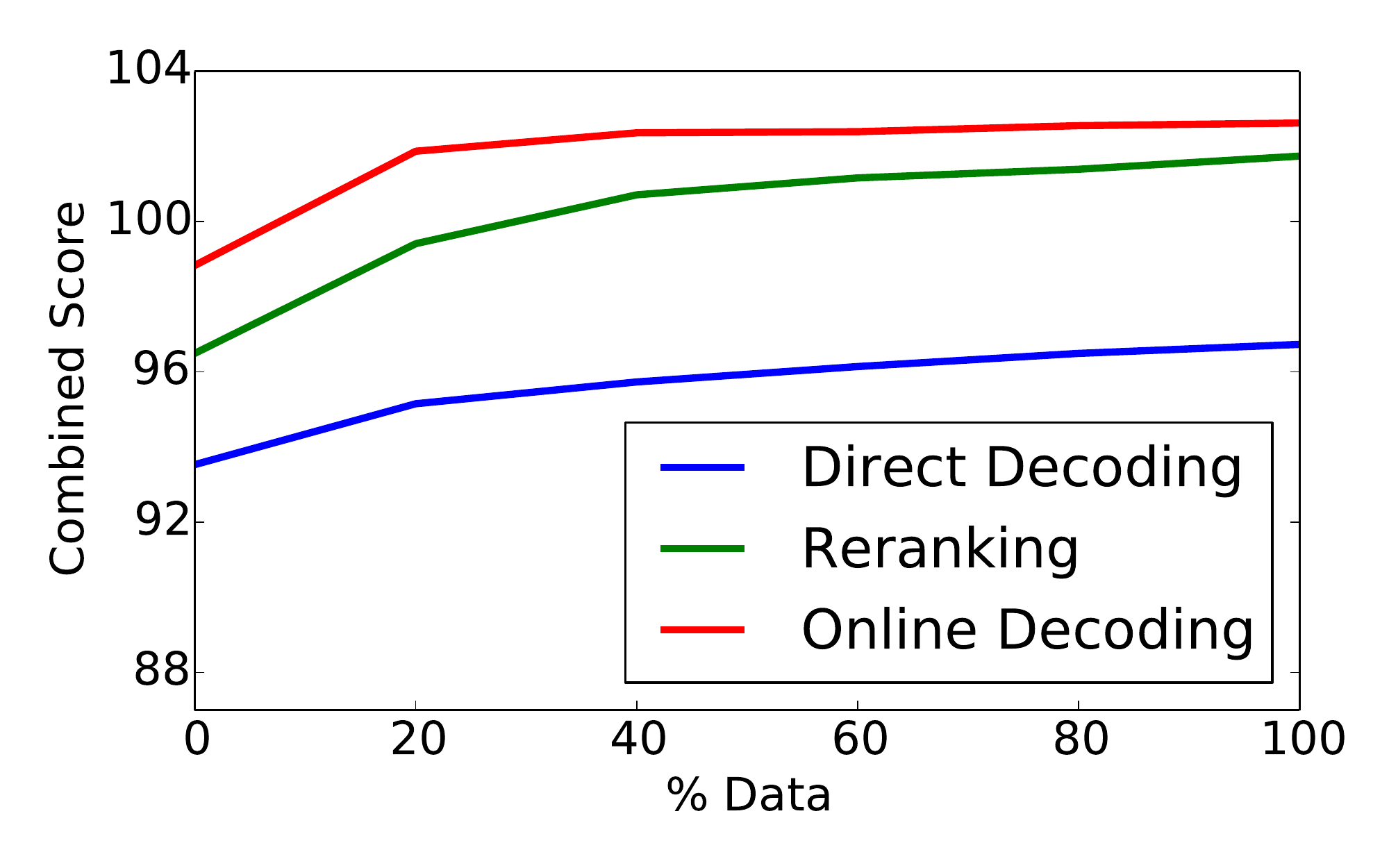}
        \vspace{-1.0em}
        \caption{Task-oriented pretraining, MultiWOZ}    
    \end{subfigure}
    \vspace{-0.5em}
    \caption{Results showing the effect of pretraining scale. \label{fig:pretrain_data}}
    \vspace{-0.5em}
\end{figure*}

\begin{figure*}[t]
    \centering 
    \begin{subfigure}[b]{0.33\textwidth}
        \centering 
        \includegraphics[height=85pt]{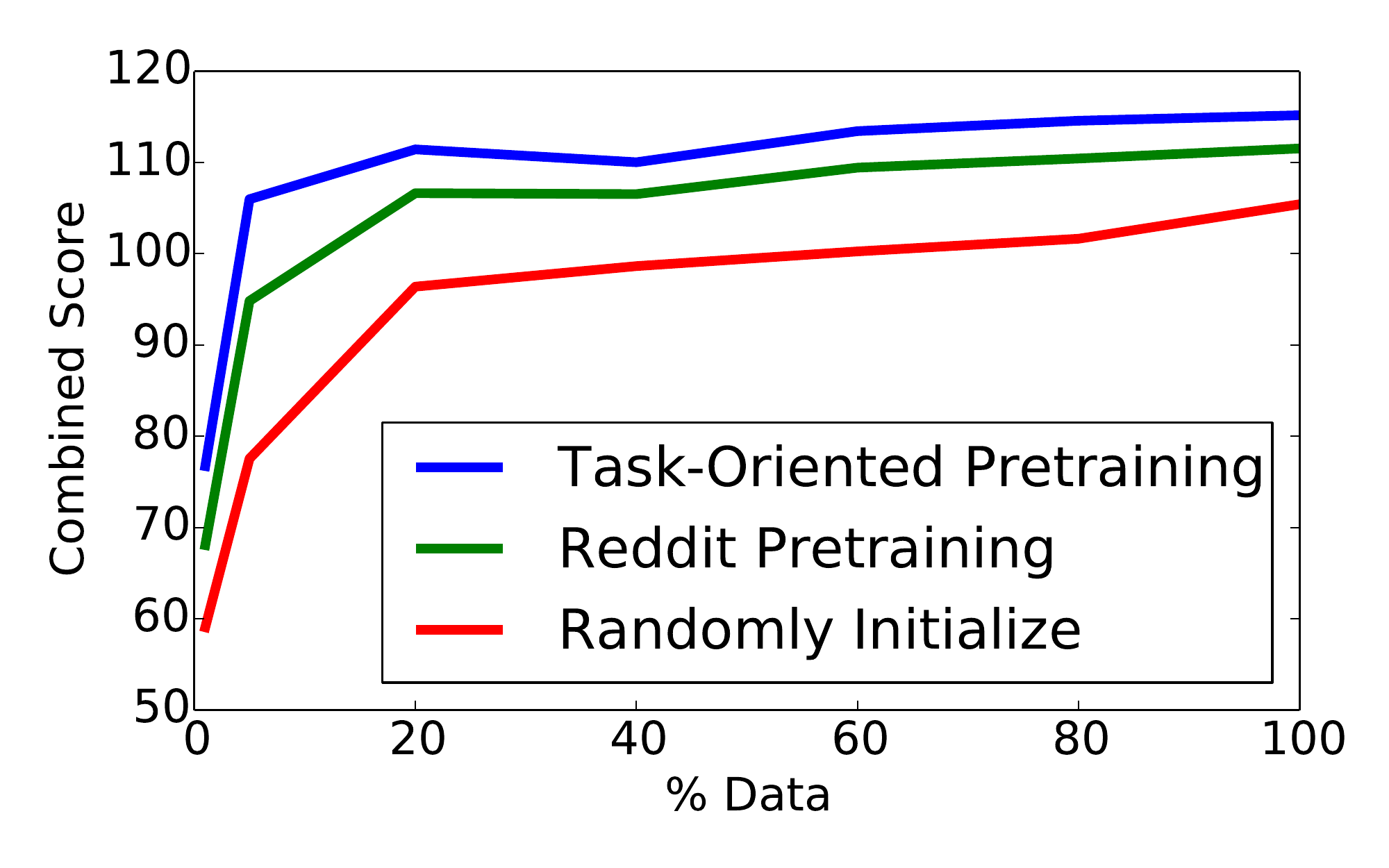}
        \vspace{-1.0em}
        \caption{Direct decoding, CamRest676}    
    \end{subfigure} \hspace{-0.5em}
    \begin{subfigure}[b]{0.33\textwidth}
        \centering 
        \includegraphics[height=85pt]{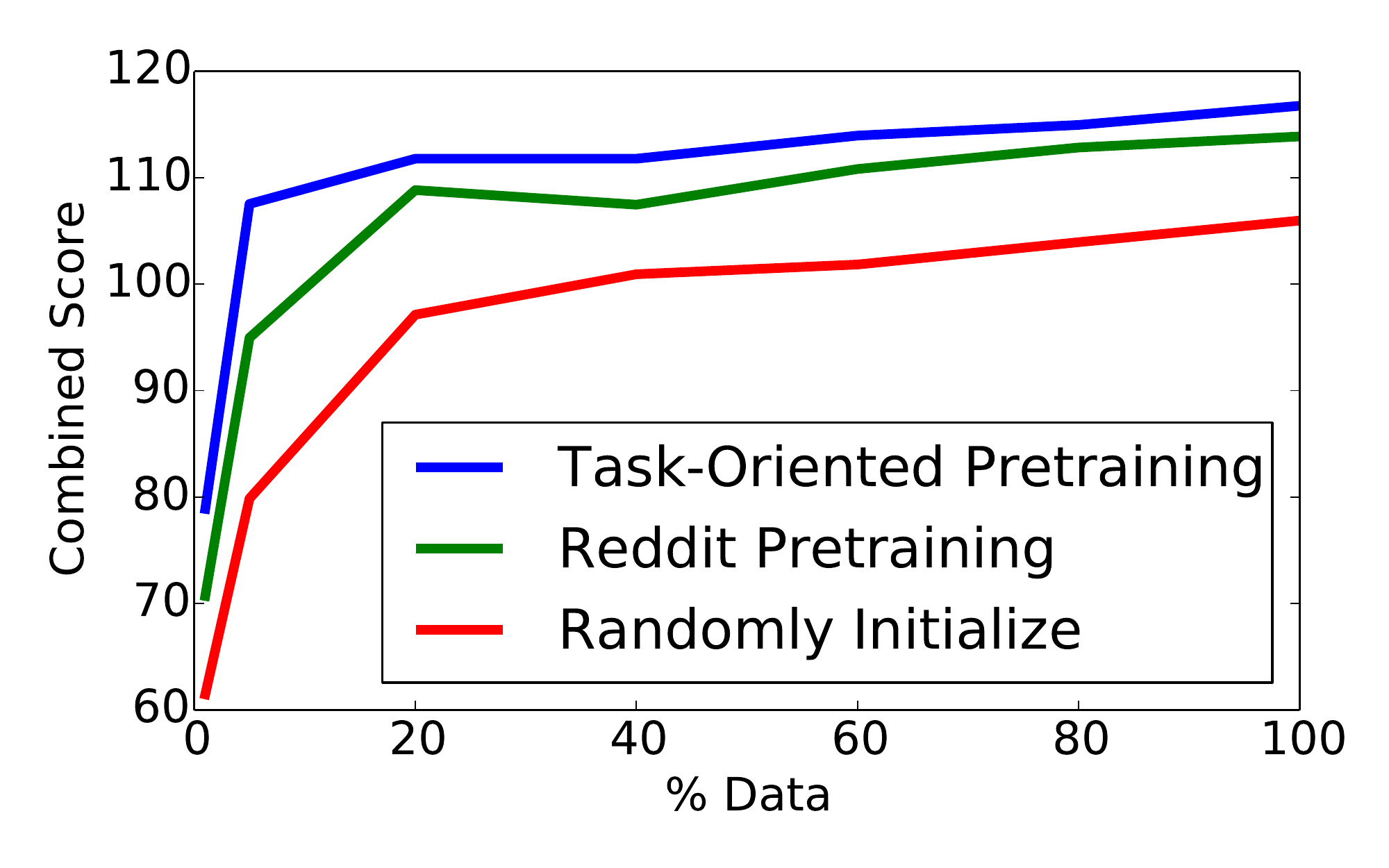}
        \vspace{-1.0em}
        \caption{Reranking, CamRest676}    
    \end{subfigure} \hspace{-0.5em}
    \begin{subfigure}[b]{0.33\textwidth}
        \centering 
        \includegraphics[height=85pt]{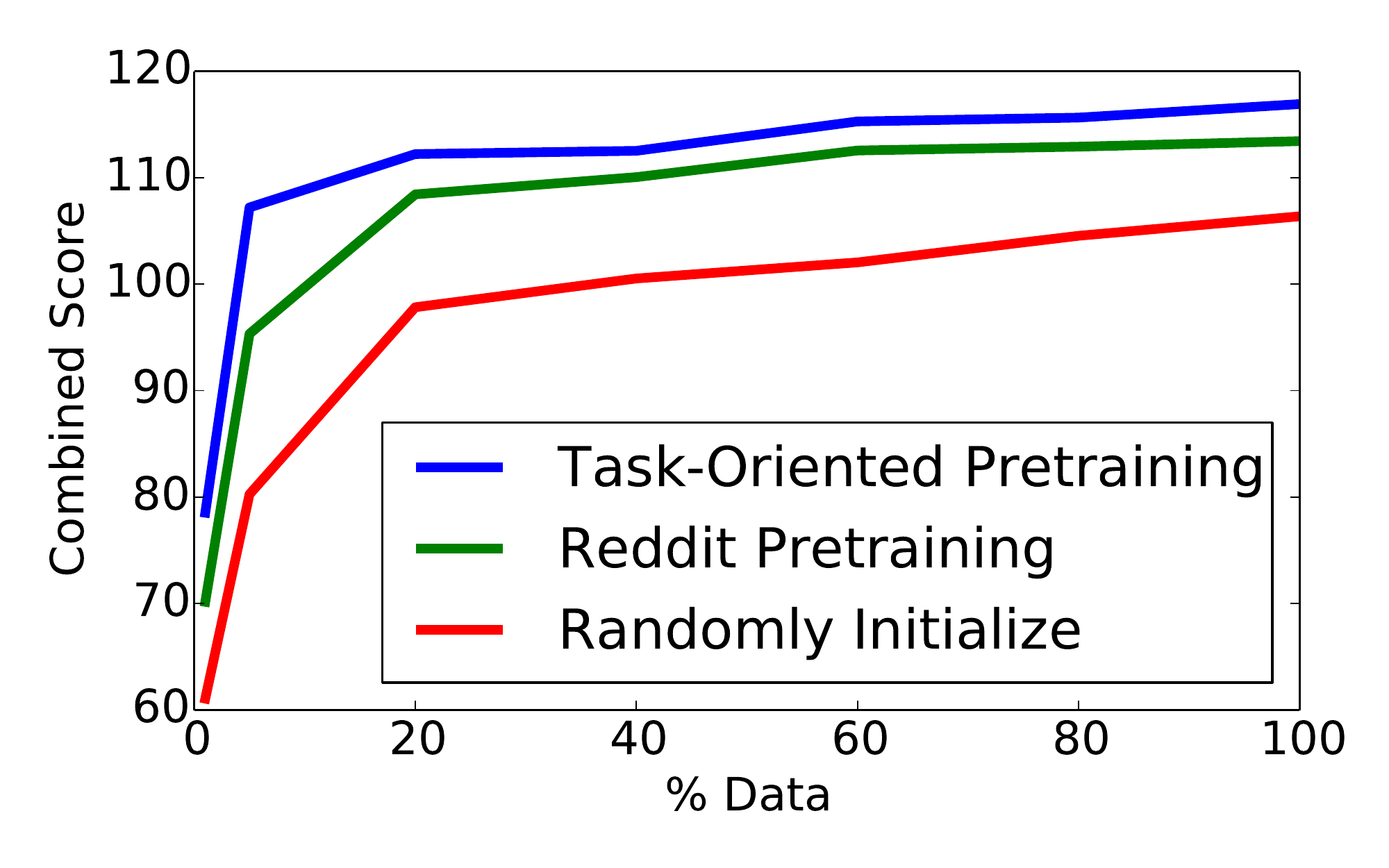}
        \vspace{-1.0em}
        \caption{Online decoding, CamRest676}
    \end{subfigure} \hspace{-0.5em}
    
    \begin{subfigure}[b]{0.33\textwidth}
        \centering 
        \includegraphics[height=85pt]{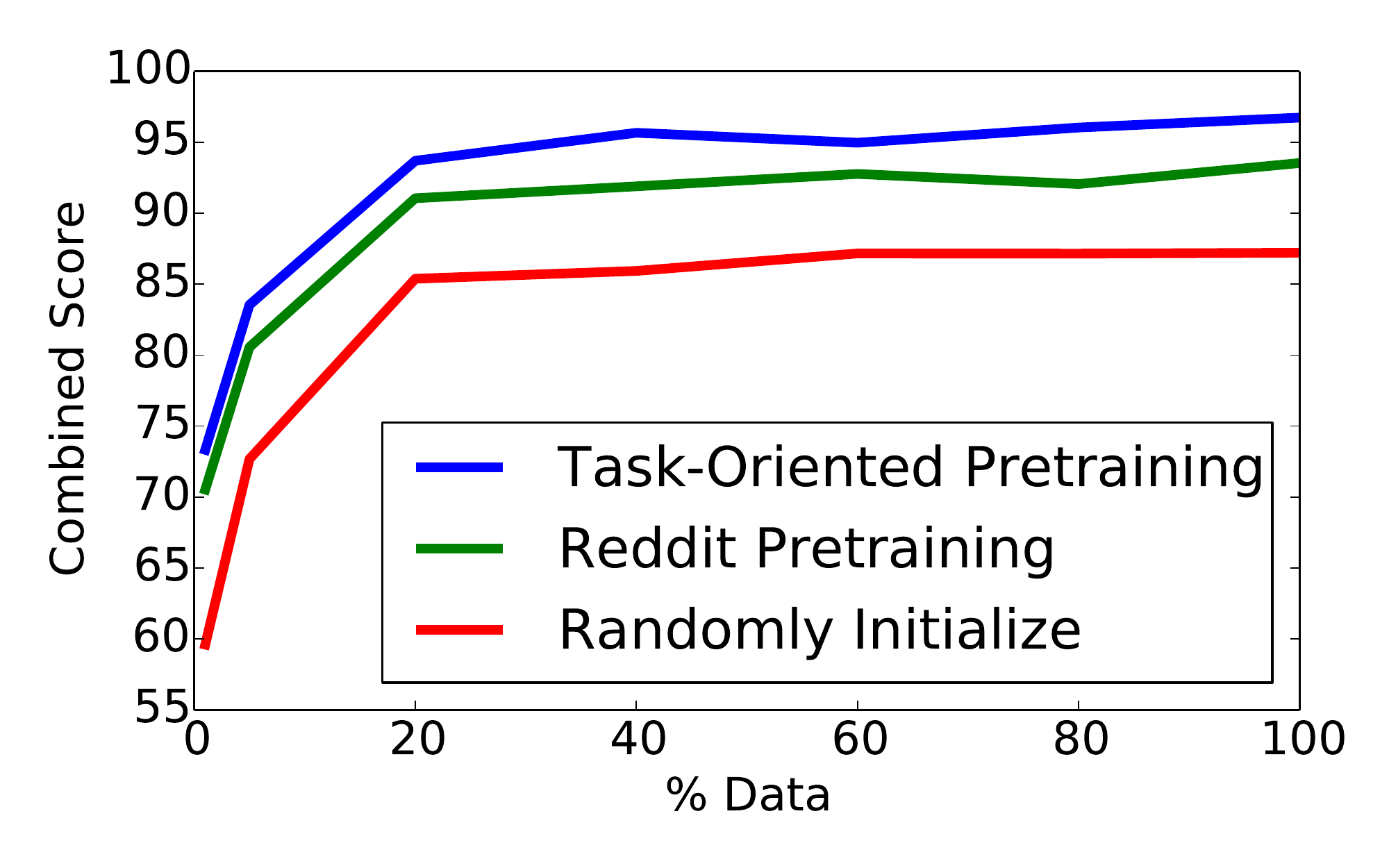}
        \vspace{-1.0em}
        \caption{Direct decoding, MultiWOZ}    
    \end{subfigure} \hspace{-0.5em}
    \begin{subfigure}[b]{0.33\textwidth}
        \centering 
        \includegraphics[height=85pt]{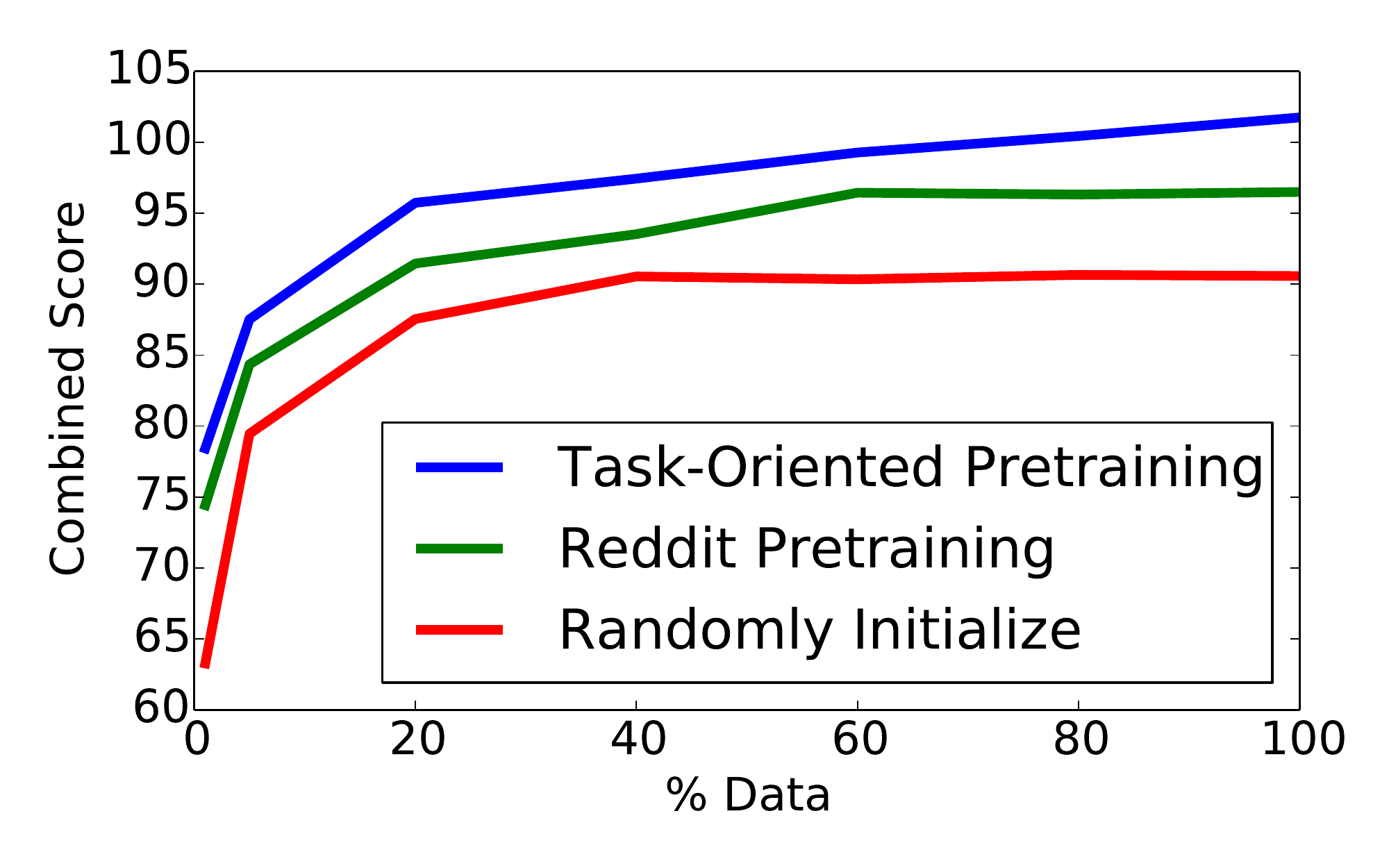}
        \vspace{-1.0em}
        \caption{Reranking, MultiWOZ}    
    \end{subfigure} \hspace{-0.5em}
    \begin{subfigure}[b]{0.33\textwidth}
        \centering 
        \includegraphics[height=85pt]{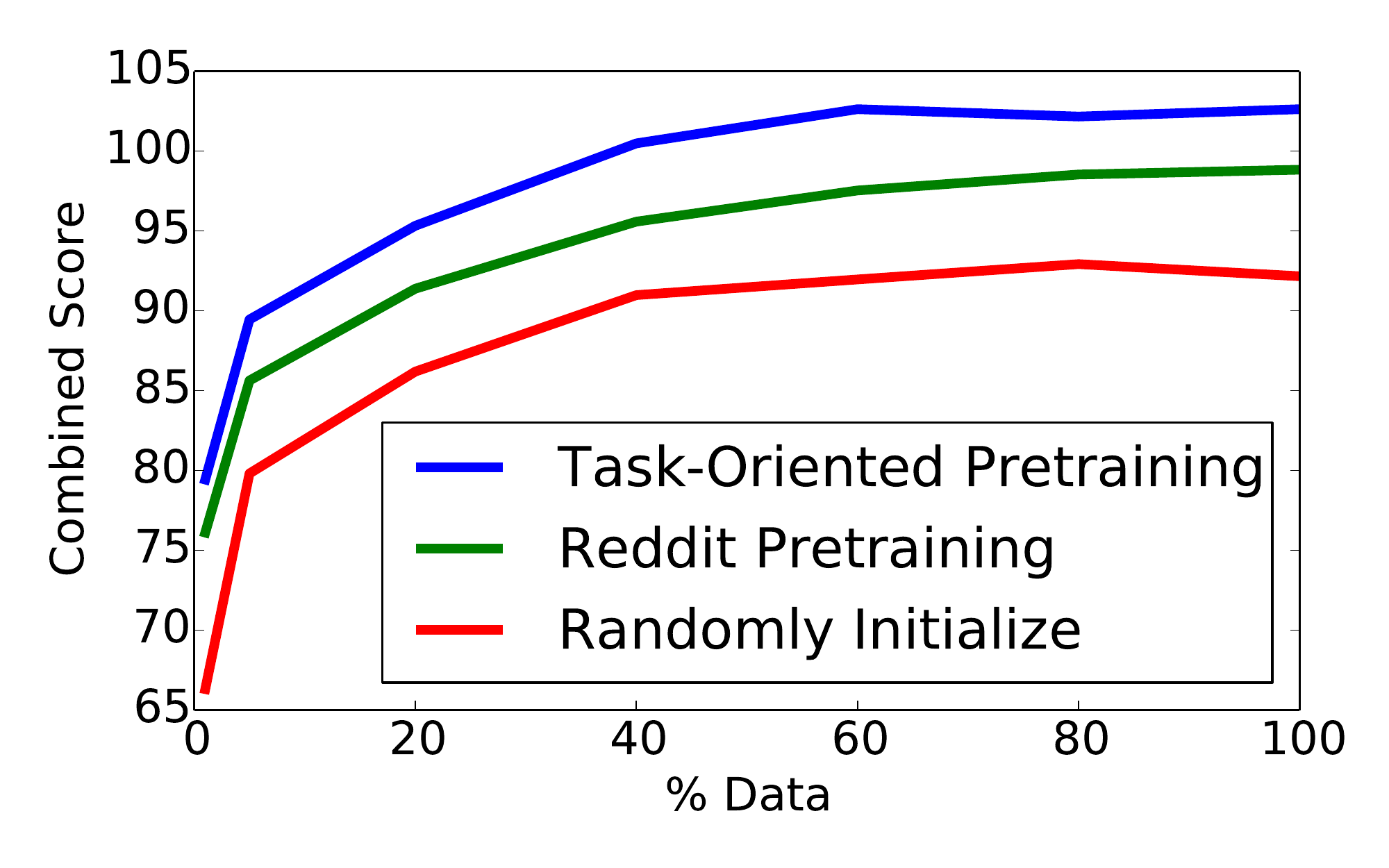}
        \vspace{-1.0em}
        \caption{Online decoding, MultiWOZ}    
    \end{subfigure}
    \vspace{-0.5em}
    \caption{Pretraining improves sample efficiency during fine-tuning.\label{fig:sample_efficiency}} 
    \vspace{-0.5em}
\end{figure*}


\begin{table}[t]
    \centering
    \footnotesize
    \renewcommand{\arraystretch}{0.9}
    \begin{tabular}{lcc}
        \toprule
        \textbf{Model}  & \textbf{CamRest676} & \textbf{MultiWOZ} \\ \midrule
        Direct decoding & \textbf{4.89} & \textbf{6.48} \\
        Reranking & 5.43 & 6.92 \\
        Online decoding & 8.73 &  10.97 \\
    \bottomrule
    \end{tabular}
    \vspace{-0.5em}
    \caption{Average decoding time (in seconds) for each turn with different decoding methods. \label{tab:runtime_analysis}}
    \vspace{-0.5em}
\end{table}

\begin{table}[t]
    \centering
    \footnotesize
    \renewcommand{\arraystretch}{0.9}
    \begin{tabular}{lcc}
        \toprule
        \textbf{Model}  & \textbf{CamRest676} & \textbf{MultiWOZ} \\ \midrule
        Ground truth & 14.50 & 16.91\\
        Direct decoding & 12.07 & 12.85\\
        Direct decoding + Length & 15.98 & 17.73\\
        Reranking & \textbf{15.09} & 17.47\\
        Online decoding & 15.14 & \textbf{17.32}\\
    \bottomrule
    \end{tabular}
    \vspace{-0.5em}
    \caption{The average length of responses with different decoding methods (on test set). The value closest to the ground truth is bold. \label{tab:avg_len}}
    \vspace{-1em}
\end{table}

\begin{table}[t]
    \centering
    \footnotesize
    \renewcommand{\arraystretch}{0.9}
    \begin{tabular}{lcc}
        \toprule
        \textbf{Model}  & \textbf{CamRest676} & \textbf{MultiWOZ} \\ \midrule
        Ground truth & 1.07 & 1.22 \\
        Direct decoding & 0.84 & 0.91\\
        Reranking & 0.87 & 0.99\\
        Online decoding & \textbf{0.89} & \textbf{1.03}\\
    \bottomrule
    \end{tabular}
    \vspace{-0.5em}
    \caption{The Zipf scores of responses with different decoding methods (on test set). The value closest to the ground truth is bold. \label{tab:zipf_score}}
    
    \vspace{0.5em}
    \centering
    \footnotesize
    \renewcommand{\arraystretch}{0.9}
    \begin{tabular}{lcc}
        \toprule
        \textbf{Model}  & \textbf{CamRest676} & \textbf{MultiWOZ} \\ \midrule
        Direct decoding & 0.24 & 0.31\\
        Reranking & 0.12 & 0.14\\
        Online decoding & \textbf{0.08} & \textbf{0.11}\\
    \bottomrule
    \end{tabular}
    \vspace{-0.5em}
    \caption{The likelihood (\%) of falling into repetition loops for different decoding methods (on test set). \label{tab:repetition_loop}}
    \vspace{-1em}
\end{table}

\subsection{Ablation on Model Combination \label{sec:model_ablation}}

Noisy channel decoding involves a combination of four sub-modules, as in Eq.\ \ref{eq:model_combination}: the direct model, channel model, language model, and length bias. We perform an ablation study to determine whether all model components are important to the result, using the large model. Results on the development sets of CamRest676 and MultiWOZ are presented in Table \ref{tab:ablation_study}. Note that the ablation is performed after applying the direct model to obtain $k_1$ expansions at each beam search step for noisy channel online decoding. We find that the combination of all four sub-modules performs the best, followed by combinations of three and then two sub-modules. The results are significant when comparing `All' and the baselines ($p < 0.01$). This result demonstrates the effectiveness of the noisy channel factorization, and the importance of each model component.


\subsection{Effect of Pretraining Scale
\label{sec:effect_pretraining}}

We investigate the importance of scale for both our pretraining stages. We select different checkpoints for Reddit pretraining, and truncate the two task-oriented dialogue datasets for task-oriented pretraining. We fine-tune these models using the full training data of CamRest676 or MultiWOZ. The results of three decoding methods (with the large noisy channel model) on the development sets are shown in Figure \ref{fig:pretrain_data}. In Figure \ref{fig:pretrain_data} (a) and (c), the combined scores of all three decoding methods improve with more Reddit pretraining steps, demonstrating the advantage of increasing amounts of data in the open-domain dialogue pretraining stage. In Figure \ref{fig:pretrain_data} (b) and (d), the combined scores further increase with more task-oriented data, confirming that additional task-oriented pretraining data is useful.

\begin{table*}[t]
    \centering
    \footnotesize
    \scalebox{0.83}{
    \begin{tabular}{l@{\hskip 0.08in}l@{\hskip 0.08in}c}
        \toprule
        \textbf{Model}   & \multicolumn{1}{c}{\textbf{Delexicalized Response}} & \textbf{SacreBLEU $\uparrow$} \\ \midrule
     \multicolumn{2}{c}{Dialogue: MUL1624, Turn Number: 4}  \\ \midrule    
    Ground truth & Sure. The trainid is  $[$train\_id$]$. It leaves at  $[$value\_time$]$ and arrives at $[$value\_time$]$. Would you like to make a reservation? & - \\
    Direct decoding & The travel time is $[$value\_count$]$ minutes. & 2.03 \\
    \multirow{2}{*}{Reranking} & \multirow{2}{*}{\parbox{15cm}{$[$train\_id$]$ leaves $[$value\_place$]$ at $[$value\_time$]$ and arrives in $[$value\_place$]$ at $[$value\_time$]$. Would you like me to book that for you?}}  & \multirow{2}{*}{46.33} \\ \\
    Online decoding & $[$train\_id$]$ leaves at $[$value\_time$]$ and arrives at $[$value\_time$]$. Would you like me to book that? & \textbf{60.90} \\ \midrule
     \multicolumn{2}{c}{Dialogue: MUL1276, Turn Number: 6}  \\ \midrule
     Ground truth & The $[$restaurant\_name$]$ has a postal code of $[$restaurant\_postcode$]$. Will this work for you? I can book it if you please. & - \\
    Direct decoding & There are $[$value\_count$]$ places. What type of food would you like? & 1.63 \\
    Reranking & How about $[$restaurant\_name$]$ located at $[$restaurant\_address$]$, postcode $[$restaurant\_postcode$]$. & \textbf{24.90} \\
    Online decoding & How about $[$restaurant\_name$]$ located at $[$restaurant\_address$]$, postcode $[$restaurant\_postcode$]$? & 22.00\\ \midrule
    
     \multicolumn{2}{c}{Dialogue: MUL1898, Turn Number: 2}  \\ \midrule
     
    \multirow{2}{*}{Ground truth} & \multirow{2}{*}{\parbox{15cm}{Yes, $[$attraction\_name$]$ is on $[$attraction\_address$]$ and is in the $[$value\_area$]$ side of town. Is there anything else you need to know?}}  & \multirow{2}{*}{-} \\ \\
    
     Direct decoding & $[$attraction\_name$]$ is located in the $[$value\_area$]$ part of town and has free admission. & 27.53 \\
    
    \multirow{2}{*}{Reranking} & \multirow{2}{*}{\parbox{15cm}{$[$attraction\_name$]$ is located in the $[$value\_area$]$ of town at $[$attraction\_address$]$. The entrance fee is free. Can I help you with anything else?}}  & \multirow{2}{*}{41.66} \\ \\
    
    Online decoding & $[$attraction\_name$]$ is located in the $[$value\_area$]$ part of town at $[$attraction\_address$]$. Can I help you with anything else? & \textbf{42.38} \\
    \bottomrule
    \end{tabular}
    }
    \caption{Case study on the responses decoded by direct decoding, noisy channel reranking and noisy channel online decoding. The large noisy channel model is used. \label{tab:case_study}}
\end{table*}

\subsection{Sample Efficiency of Fine-Tuning \label{sec:sample_efficiency}}

We investigate whether pretraining can improve sample efficiency during fine-tuning. We gradually increase the amount of fine-tuning data and evaluate the randomly-initialized, Reddit pretrained and task-oriented pretrained models.
The results on the development sets are shown in Figure \ref{fig:sample_efficiency}. Combined scores increase with more training data under all conditions. Crucially, Reddit pretrained models show better performance with a smaller amount of fine-tuning data than randomly initialized models, and task-oriented pretrained models better still. We conclude that both our pretraining stages can improve sample efficiency, which is especially important when the target task has little training data.


\subsection{Decoding Runtime \label{sec:runtime_analysis}}

In Table \ref{tab:runtime_analysis}, we report the average clock time for decoding one turn (including its belief state, dialogue act and response). Noisy channel reranking is slightly slower compared to direct decoding, with overhead due to the reranking step in Eq.\ \ref{eq:model_combination}. Noisy channel online decoding is significantly slower, since it needs to apply Eq.\ \ref{eq:model_combination} at each beam search step. In future work we will investigate ways to improve the efficiency of online decoding.

\subsection{Decoding Properties \label{sec:decoded_response_analysis}}

In this section we analyze why the noisy channel model performed better than direct decoding.

\textbf{Length}: In Table \ref{tab:avg_len} we show the average length of generated responses. Direct decoding produces shorter responses than the ground truth, 
confirming that the direct model prefers short and generic responses.
Adding a length bias to direct decoding (with lambda tuned on the development sets) produces responses longer than the ground truth, which may be a disadvantage. The noisy channel models produce responses with average length closest to the ground truth.

\textbf{Zipf}: Table \ref{tab:zipf_score} shows the Zipf scores of responses. We find that the word distributions of responses generated by the noisy channel models are closer to the word distribution of ground-truth responses.

\textbf{Repetition}: In Table \ref{tab:repetition_loop} we examine the likelihood of falling into repetition loops \cite{holtzman2019curious} for different decoding methods. Repetition loops are rare for all decoding methods, but noisy channel decoding can further decrease their likelihood. The channel model can discount a sequence with a repetition loop, since it conveys less information than a natural sequence of the same length, making it harder to ``explain'' the context.

\textbf{Examples}: Some examples of responses are shown in Table \ref{tab:case_study}. We observe that noisy channel models decode longer responses compared to direct decoding, and that the responses can explain their dialogue contexts well to meet users' requirements.

\section{Related Work}

\textbf{Task-oriented dialogue models}: Most task-oriented dialogue systems break down the task into three components: belief tracking \cite{henderson2013deep,mrkvsic2016neural,rastogi2017scalable,nouri2018toward,wu2019transferable,zhang2019find,zhou2019multi,heck2020trippy}, dialogue act prediction \cite{wen2017latent,tanaka2019dialogue} and response generation \cite{wen2015semantically,budzianowski2018multiwoz,lippe2020diversifying}. Traditionally, a modular approach is adopted, where these components are optimized independently (i.e.\ a pipeline design) or learned via multi-task learning (i.e.\ some parameters are shared among the components) \cite{wen2016network,neelakantan2019neural,zhao2019rethinking,mehri2019structured,tseng2020generative,lee2020sumbt}. However, it is known that improvements in one component do not necessarily lead to overall performance improvements \cite{ham2020end}, and the modular approach suffers from error propagation in practice \cite{liu2018end}. These observations gave rise to the sequence-to-sequence approach \cite{lei2018sequicity,pei2019modular,budzianowski2019hello,wu2019alternating,zhang2019task,ham2020end,hosseini2020simple,peng2020soloist,yang2020ubar}, where dialogue beliefs and acts are represented as text spans, and a sequence-to-sequence model is applied to subsume the three components. Our work is situated within this general approach. In contrast to previous work, however, which uses a direct model for decoding, we introduce the noisy channel model to improve task-oriented dialogue.

\textbf{Pretraining models for dialogue}: Recent work has applied pretraining \cite{peters2018deep,devlin2018bert,radford2019language} to dialogue. For open-domain dialogue, DialoGPT \cite{zhang2019dialogpt} and CGRG \cite{wu2020controllable} extend GPT-2 \cite{radford2019language} for response generation. PLATO \cite{bao2019plato} and PLATO-2 \cite{bao2020plato} pretrain a latent variable model with social media data for diversified response generation. Meena \cite{adiwardana2020towards} collects a large-scale social media corpus for pretraining and proposes a metric named sensibleness and specificity average for evaluation. Roller et al.\ \shortcite{roller2020recipes} study various strategies for building an open-domain chatbot with Reddit for pretraining. For task-oriented dialogue, ToD-BERT \cite{wu2020tod} fine-tunes BERT \cite{devlin2018bert} for four tasks, including intention detection, belief tracking, dialogue act prediction, and response selection. SC-GPT \cite{peng2020few} fine-tunes GPT-2 for few-shot response generation with given dialogue acts. Ham et al.\ \shortcite{ham2020end} fine-tune GPT-2 for belief tracking and context-to-response generation. SimpleTOD \cite{hosseini2020simple} proposes a method to serialize dialogue beliefs and acts into text spans and fine-tunes GPT-2 for end-to-end dialogue modeling. SOLOIST \cite{peng2020soloist} uses a series of task-oriented dialogue datasets to further pretrain GPT-2 before fine-tuning it on final tasks for evaluation. Unlike these BERT- or GPT-initialized task-oriented dialogue models, which are essentially pretrained with general text, such as Wikipedia and BookCorpus, we use a Reddit dump to pretrain the models to learn from open-domain dialogues. 

\section*{Conclusion}
We introduced two noisy channel models, noisy channel reranking and noisy channel online decoding, for task-oriented dialogue. Large-scale pretraining was further adopted to tackle data scarcity in downstream tasks. Extensive experiments on MultiWOZ, CamRest676 and SMCalFlow demonstrated that (1) the noisy channel models significantly outperform direct decoding; (2) models with pretraining improve over randomly-initialized models; (3) the models are robust to different dialogue schema annotations; (4) the noisy channel models can decode responses closer to ground-truth responses than direct decoding.

\section*{Acknowledgements}
We would like to thank the action editors (Maggie, Wenjie Li and Eneko Agirre) and three anonymous reviewers for their insightful comments. We also thank Angeliki Lazaridou, Gábor Melis, Nando de Freitas, Chris Dyer and the DeepMind language team for their helpful discussions.

\bibliography{tacl2018}

\begin{thebibliography}{67}
\expandafter\ifx\csname natexlab\endcsname\relax\def\natexlab#1{#1}\fi

\bibitem[{Adiwardana et~al.(2020)Adiwardana, Luong, So, Hall, Fiedel,
  Thoppilan, Yang, Kulshreshtha, Nemade, Lu et~al.}]{adiwardana2020towards}
Daniel Adiwardana, Minh-Thang Luong, David~R So, Jamie Hall, Noah Fiedel, Romal
  Thoppilan, Zi~Yang, Apoorv Kulshreshtha, Gaurav Nemade, Yifeng Lu, et~al.
  2020.
\newblock Towards a human-like open-domain chatbot.
\newblock \emph{arXiv preprint arXiv:2001.09977}.

\bibitem[{Andreas et~al.(2020)Andreas, Bufe, Burkett, Chen, Clausman, Crawford,
  Crim, DeLoach, Dorner, Eisner et~al.}]{andreas2020task}
Jacob Andreas, John Bufe, David Burkett, Charles Chen, Josh Clausman, Jean
  Crawford, Kate Crim, Jordan DeLoach, Leah Dorner, Jason Eisner, et~al. 2020.
\newblock Task-oriented dialogue as dataflow synthesis.
\newblock \emph{Transactions of the Association for Computational Linguistics},
  8:556--571.

\bibitem[{Austin(1975)}]{austin1975things}
John~Langshaw Austin. 1975.
\newblock \emph{How to do things with words}, volume~88.
\newblock Oxford university press.

\bibitem[{Bao et~al.(2019)Bao, He, Wang, and Wu}]{bao2019plato}
Siqi Bao, Huang He, Fan Wang, and Hua Wu. 2019.
\newblock Plato: Pre-trained dialogue generation model with discrete latent
  variable.
\newblock \emph{arXiv preprint arXiv:1910.07931}.

\bibitem[{Bao et~al.(2020)Bao, He, Wang, Wu, and Wang}]{bao2020plato}
Siqi Bao, Huang He, Fan Wang, Hua Wu, and Haifeng Wang. 2020.
\newblock \href {https://doi.org/10.18653/v1/2020.acl-main.9} {{PLATO:}
  pre-trained dialogue generation model with discrete latent variable}.
\newblock In \emph{Proceedings of the 58th Annual Meeting of the Association
  for Computational Linguistics, {ACL} 2020, Online, July 5-10, 2020}, pages
  85--96. Association for Computational Linguistics.

\bibitem[{Bradbury et~al.(2018)Bradbury, Frostig, Hawkins, Johnson, Leary,
  Maclaurin, and Wanderman-Milne}]{jax2018github}
James Bradbury, Roy Frostig, Peter Hawkins, Matthew~James Johnson, Chris Leary,
  Dougal Maclaurin, and Skye Wanderman-Milne. 2018.
\newblock \href {http://github.com/google/jax} {{JAX}: composable
  transformations of {P}ython+{N}um{P}y programs}.

\bibitem[{Brants et~al.(2007)Brants, Popat, Xu, Och, and
  Dean}]{brants2007large}
Thorsten Brants, Ashok~C. Popat, Peng Xu, Franz~J. Och, and Jeffrey Dean. 2007.
\newblock Large language models in machine translation.
\newblock In \emph{Proceedings of the 2007 Joint Conference on Empirical
  Methods in Natural Language Processing and Computational Natural Language
  Learning ({EMNLP}-{C}o{NLL})}, pages 858--867, Prague, Czech Republic.
  Association for Computational Linguistics.

\bibitem[{Brown et~al.(2020)Brown, Mann, Ryder, Subbiah, Kaplan, Dhariwal,
  Neelakantan, Shyam, Sastry, Askell, Agarwal, Herbert{-}Voss, Krueger,
  Henighan, Child, Ramesh, Ziegler, Wu, Winter, Hesse, Chen, Sigler, Litwin,
  Gray, Chess, Clark, Berner, McCandlish, Radford, Sutskever, and
  Amodei}]{brown2020language}
Tom~B. Brown, Benjamin Mann, Nick Ryder, Melanie Subbiah, Jared Kaplan,
  Prafulla Dhariwal, Arvind Neelakantan, Pranav Shyam, Girish Sastry, Amanda
  Askell, Sandhini Agarwal, Ariel Herbert{-}Voss, Gretchen Krueger, Tom
  Henighan, Rewon Child, Aditya Ramesh, Daniel~M. Ziegler, Jeffrey Wu, Clemens
  Winter, Christopher Hesse, Mark Chen, Eric Sigler, Mateusz Litwin, Scott
  Gray, Benjamin Chess, Jack Clark, Christopher Berner, Sam McCandlish, Alec
  Radford, Ilya Sutskever, and Dario Amodei. 2020.
\newblock \href
  {https://proceedings.neurips.cc/paper/2020/hash/1457c0d6bfcb4967418bfb8ac142f64a-Abstract.html}
  {Language models are few-shot learners}.
\newblock In \emph{Advances in Neural Information Processing Systems 33: Annual
  Conference on Neural Information Processing Systems 2020, NeurIPS 2020,
  December 6-12, 2020, virtual}.

\bibitem[{Budzianowski and Vuli{\'c}(2019)}]{budzianowski2019hello}
Pawe{\l} Budzianowski and Ivan Vuli{\'c}. 2019.
\newblock \href {https://doi.org/10.18653/v1/D19-5602} {Hello, it{'}s {GPT}-2 -
  how can {I} help you? towards the use of pretrained language models for
  task-oriented dialogue systems}.
\newblock In \emph{Proceedings of the 3rd Workshop on Neural Generation and
  Translation}, pages 15--22, Hong Kong. Association for Computational
  Linguistics.

\bibitem[{Budzianowski et~al.(2018)Budzianowski, Wen, Tseng, Casanueva, Ultes,
  Ramadan, and Gasic}]{budzianowski2018multiwoz}
Pawel Budzianowski, Tsung{-}Hsien Wen, Bo{-}Hsiang Tseng, I{\~{n}}igo
  Casanueva, Stefan Ultes, Osman Ramadan, and Milica Gasic. 2018.
\newblock Multiwoz - {A} large-scale multi-domain wizard-of-oz dataset for
  task-oriented dialogue modelling.
\newblock In \emph{Proceedings of the 2018 Conference on Empirical Methods in
  Natural Language Processing, Brussels, Belgium, October 31 - November 4,
  2018}, pages 5016--5026. Association for Computational Linguistics.

\bibitem[{Byrne et~al.(2019)Byrne, Krishnamoorthi, Sankar, Neelakantan,
  Goodrich, Duckworth, Yavuz, Dubey, Kim, and Cedilnik}]{byrne2019taskmaster}
Bill Byrne, Karthik Krishnamoorthi, Chinnadhurai Sankar, Arvind Neelakantan,
  Ben Goodrich, Daniel Duckworth, Semih Yavuz, Amit Dubey, Kyu{-}Young Kim, and
  Andy Cedilnik. 2019.
\newblock \href {https://doi.org/10.18653/v1/D19-1459} {Taskmaster-1: Toward a
  realistic and diverse dialog dataset}.
\newblock In \emph{Proceedings of the 2019 Conference on Empirical Methods in
  Natural Language Processing and the 9th International Joint Conference on
  Natural Language Processing, {EMNLP-IJCNLP} 2019, Hong Kong, China, November
  3-7, 2019}, pages 4515--4524. Association for Computational Linguistics.

\bibitem[{Chen et~al.(2019)Chen, Chen, Qin, Yan, and
  Wang}]{wen2015semantically}
Wenhu Chen, Jianshu Chen, Pengda Qin, Xifeng Yan, and William~Yang Wang. 2019.
\newblock \href {https://doi.org/10.18653/v1/P19-1360} {Semantically
  conditioned dialog response generation via hierarchical disentangled
  self-attention}.
\newblock In \emph{Proceedings of the 57th Annual Meeting of the Association
  for Computational Linguistics}, pages 3696--3709, Florence, Italy.
  Association for Computational Linguistics.

\bibitem[{Devlin et~al.(2019)Devlin, Chang, Lee, and
  Toutanova}]{devlin2018bert}
Jacob Devlin, Ming{-}Wei Chang, Kenton Lee, and Kristina Toutanova. 2019.
\newblock \href {https://doi.org/10.18653/v1/n19-1423} {{BERT:} pre-training of
  deep bidirectional transformers for language understanding}.
\newblock In \emph{Proceedings of the 2019 Conference of the North American
  Chapter of the Association for Computational Linguistics: Human Language
  Technologies, {NAACL-HLT} 2019, Minneapolis, MN, USA, June 2-7, 2019, Volume
  1 (Long and Short Papers)}, pages 4171--4186. Association for Computational
  Linguistics.

\bibitem[{Fan et~al.(2018)Fan, Lewis, and Dauphin}]{fan2018hierarchical}
Angela Fan, Mike Lewis, and Yann~N. Dauphin. 2018.
\newblock \href {https://doi.org/10.18653/v1/P18-1082} {Hierarchical neural
  story generation}.
\newblock In \emph{Proceedings of the 56th Annual Meeting of the Association
  for Computational Linguistics, {ACL} 2018, Melbourne, Australia, July 15-20,
  2018, Volume 1: Long Papers}, pages 889--898. Association for Computational
  Linguistics.

\bibitem[{Ham et~al.(2020)Ham, Lee, Jang, and Kim}]{ham2020end}
Donghoon Ham, Jeong-Gwan Lee, Youngsoo Jang, and Kee-Eung Kim. 2020.
\newblock \href {https://doi.org/10.18653/v1/2020.acl-main.54} {End-to-end
  neural pipeline for goal-oriented dialogue systems using {GPT}-2}.
\newblock In \emph{Proceedings of the 58th Annual Meeting of the Association
  for Computational Linguistics}, pages 583--592, Online. Association for
  Computational Linguistics.

\bibitem[{Heck et~al.(2020)Heck, van Niekerk, Lubis, Geishauser, Lin, Moresi,
  and Gasic}]{heck2020trippy}
Michael Heck, Carel van Niekerk, Nurul Lubis, Christian Geishauser,
  Hsien{-}Chin Lin, Marco Moresi, and Milica Gasic. 2020.
\newblock \href {https://www.aclweb.org/anthology/2020.sigdial-1.4/} {Trippy:
  {A} triple copy strategy for value independent neural dialog state tracking}.
\newblock In \emph{Proceedings of the 21th Annual Meeting of the Special
  Interest Group on Discourse and Dialogue, SIGdial 2020, 1st virtual meeting,
  July 1-3, 2020}, pages 35--44. Association for Computational Linguistics.

\bibitem[{Henderson et~al.(2013)Henderson, Thomson, and
  Young}]{henderson2013deep}
Matthew Henderson, Blaise Thomson, and Steve Young. 2013.
\newblock Deep neural network approach for the dialog state tracking challenge.
\newblock In \emph{Proceedings of the SIGDIAL 2013 Conference}, pages 467--471.

\bibitem[{Henderson et~al.(2019)Henderson, Vulic, Gerz, Casanueva,
  Budzianowski, Coope, Spithourakis, Wen, Mrksic, and
  Su}]{henderson2019training}
Matthew Henderson, Ivan Vulic, Daniela Gerz, I{\~{n}}igo Casanueva, Pawel
  Budzianowski, Sam Coope, Georgios Spithourakis, Tsung{-}Hsien Wen, Nikola
  Mrksic, and Pei{-}Hao Su. 2019.
\newblock \href {https://doi.org/10.18653/v1/p19-1536} {Training neural
  response selection for task-oriented dialogue systems}.
\newblock In \emph{Proceedings of the 57th Conference of the Association for
  Computational Linguistics, {ACL} 2019, Florence, Italy, July 28- August 2,
  2019, Volume 1: Long Papers}, pages 5392--5404. Association for Computational
  Linguistics.

\bibitem[{Hendrycks and Gimpel(2016)}]{hendrycks2016gaussian}
Dan Hendrycks and Kevin Gimpel. 2016.
\newblock Gaussian error linear units (gelus).
\newblock \emph{arXiv preprint arXiv:1606.08415}.

\bibitem[{Hennigan et~al.(2020)Hennigan, Cai, Norman, and
  Babuschkin}]{haiku2020github}
Tom Hennigan, Trevor Cai, Tamara Norman, and Igor Babuschkin. 2020.
\newblock \href {http://github.com/deepmind/dm-haiku} {{H}aiku: {S}onnet for
  {JAX}}.

\bibitem[{Holtzman et~al.(2019)Holtzman, Buys, Forbes, and
  Choi}]{holtzman2019curious}
Ari Holtzman, Jan Buys, Maxwell Forbes, and Yejin Choi. 2019.
\newblock \href {http://arxiv.org/abs/1904.09751} {The curious case of neural
  text degeneration}.
\newblock \emph{CoRR}, abs/1904.09751.

\bibitem[{Hosseini{-}Asl et~al.(2020)Hosseini{-}Asl, McCann, Wu, Yavuz, and
  Socher}]{hosseini2020simple}
Ehsan Hosseini{-}Asl, Bryan McCann, Chien{-}Sheng Wu, Semih Yavuz, and Richard
  Socher. 2020.
\newblock \href
  {https://proceedings.neurips.cc/paper/2020/hash/e946209592563be0f01c844ab2170f0c-Abstract.html}
  {A simple language model for task-oriented dialogue}.
\newblock In \emph{Advances in Neural Information Processing Systems 33: Annual
  Conference on Neural Information Processing Systems 2020, NeurIPS 2020,
  December 6-12, 2020, virtual}.

\bibitem[{Kingma and Ba(2015)}]{kingma2014adam}
Diederik~P. Kingma and Jimmy Ba. 2015.
\newblock \href {http://arxiv.org/abs/1412.6980} {Adam: {A} method for
  stochastic optimization}.
\newblock In \emph{3rd International Conference on Learning Representations,
  {ICLR} 2015, San Diego, CA, USA, May 7-9, 2015, Conference Track
  Proceedings}.

\bibitem[{Klein and Manning(2002)}]{klein2002conditional}
Dan Klein and Christopher~D Manning. 2002.
\newblock Conditional structure versus conditional estimation in nlp models.
\newblock In \emph{Proceedings of the 2002 Conference on Empirical Methods in
  Natural Language Processing (EMNLP 2002)}, pages 9--16.

\bibitem[{Lan et~al.(2020)Lan, Chen, Goodman, Gimpel, Sharma, and
  Soricut}]{lan2019albert}
Zhenzhong Lan, Mingda Chen, Sebastian Goodman, Kevin Gimpel, Piyush Sharma, and
  Radu Soricut. 2020.
\newblock \href {https://openreview.net/forum?id=H1eA7AEtvS} {{ALBERT:} {A}
  lite {BERT} for self-supervised learning of language representations}.
\newblock In \emph{8th International Conference on Learning Representations,
  {ICLR} 2020, Addis Ababa, Ethiopia, April 26-30, 2020}.

\bibitem[{Lee et~al.(2020)Lee, Jo, Kim, Jung, and Kim}]{lee2020sumbt}
Hwaran Lee, Seokhwan Jo, HyungJun Kim, Sangkeun Jung, and Tae-Yoon Kim. 2020.
\newblock Sumbt+ larl: End-to-end neural task-oriented dialog system with
  reinforcement learning.
\newblock \emph{arXiv preprint arXiv:2009.10447}.

\bibitem[{Lei et~al.(2018)Lei, Jin, Kan, Ren, He, and Yin}]{lei2018sequicity}
Wenqiang Lei, Xisen Jin, Min-Yen Kan, Zhaochun Ren, Xiangnan He, and Dawei Yin.
  2018.
\newblock Sequicity: Simplifying task-oriented dialogue systems with single
  sequence-to-sequence architectures.
\newblock In \emph{Proceedings of the 56th Annual Meeting of the Association
  for Computational Linguistics (Volume 1: Long Papers)}, pages 1437--1447.

\bibitem[{Li et~al.(2016)Li, Galley, Brockett, Gao, and
  Dolan}]{li2015diversity}
Jiwei Li, Michel Galley, Chris Brockett, Jianfeng Gao, and Bill Dolan. 2016.
\newblock \href {https://doi.org/10.18653/v1/n16-1014} {A diversity-promoting
  objective function for neural conversation models}.
\newblock In \emph{{NAACL} {HLT} 2016, The 2016 Conference of the North
  American Chapter of the Association for Computational Linguistics: Human
  Language Technologies, San Diego California, USA, June 12-17, 2016}, pages
  110--119. The Association for Computational Linguistics.

\bibitem[{Lippe et~al.(2020)Lippe, Ren, Haned, Voorn, and
  de~Rijke}]{lippe2020diversifying}
Phillip Lippe, Pengjie Ren, Hinda Haned, Bart Voorn, and Maarten de~Rijke.
  2020.
\newblock \href {http://arxiv.org/abs/2008.03391} {Diversifying task-oriented
  dialogue response generation with prototype guided paraphrasing}.
\newblock \emph{CoRR}, abs/2008.03391.

\bibitem[{Liu and Lane(2018)}]{liu2018end}
Bing Liu and Ian Lane. 2018.
\newblock End-to-end learning of task-oriented dialogs.
\newblock In \emph{Proceedings of the 2018 Conference of the North American
  Chapter of the Association for Computational Linguistics: Student Research
  Workshop}, pages 67--73.

\bibitem[{Mehri et~al.(2019)Mehri, Srinivasan, and
  Eskenazi}]{mehri2019structured}
Shikib Mehri, Tejas Srinivasan, and Maxine Eskenazi. 2019.
\newblock Structured fusion networks for dialog.
\newblock \emph{arXiv preprint arXiv:1907.10016}.

\bibitem[{Mrk{\v{s}}i{\'c} et~al.(2016)Mrk{\v{s}}i{\'c}, S{\'e}aghdha, Wen,
  Thomson, and Young}]{mrkvsic2016neural}
Nikola Mrk{\v{s}}i{\'c}, Diarmuid~O S{\'e}aghdha, Tsung-Hsien Wen, Blaise
  Thomson, and Steve Young. 2016.
\newblock Neural belief tracker: Data-driven dialogue state tracking.
\newblock \emph{arXiv preprint arXiv:1606.03777}.

\bibitem[{Neelakantan et~al.(2019)Neelakantan, Yavuz, Narang, Prasad, Goodrich,
  Duckworth, Sankar, and Yan}]{neelakantan2019neural}
Arvind Neelakantan, Semih Yavuz, Sharan Narang, Vishaal Prasad, Ben Goodrich,
  Daniel Duckworth, Chinnadhurai Sankar, and Xifeng Yan. 2019.
\newblock Neural assistant: Joint action prediction, response generation, and
  latent knowledge reasoning.
\newblock \emph{arXiv preprint arXiv:1910.14613}.

\bibitem[{Nguyen and Salazar(2019)}]{nguyen2019transformers}
Toan~Q Nguyen and Julian Salazar. 2019.
\newblock Transformers without tears: Improving the normalization of
  self-attention.
\newblock \emph{arXiv preprint arXiv:1910.05895}.

\bibitem[{Nouri and Hosseini-Asl(2018)}]{nouri2018toward}
Elnaz Nouri and Ehsan Hosseini-Asl. 2018.
\newblock Toward scalable neural dialogue state tracking model.
\newblock \emph{arXiv preprint arXiv:1812.00899}.

\bibitem[{Pei et~al.(2019)Pei, Ren, and de~Rijke}]{pei2019modular}
Jiahuan Pei, Pengjie Ren, and Maarten de~Rijke. 2019.
\newblock A modular task-oriented dialogue system using a neural
  mixture-of-experts.
\newblock \emph{arXiv preprint arXiv:1907.05346}.

\bibitem[{Peng et~al.(2020{\natexlab{a}})Peng, Li, Li, Shayandeh, Liden, and
  Gao}]{peng2020soloist}
Baolin Peng, Chunyuan Li, Jinchao Li, Shahin Shayandeh, Lars Liden, and
  Jianfeng Gao. 2020{\natexlab{a}}.
\newblock Soloist: Few-shot task-oriented dialog with a single pre-trained
  auto-regressive model.
\newblock \emph{arXiv preprint arXiv:2005.05298}.

\bibitem[{Peng et~al.(2020{\natexlab{b}})Peng, Zhu, Li, Li, Li, Zeng, and
  Gao}]{peng2020few}
Baolin Peng, Chenguang Zhu, Chunyuan Li, Xiujun Li, Jinchao Li, Michael Zeng,
  and Jianfeng Gao. 2020{\natexlab{b}}.
\newblock \href {https://doi.org/10.18653/v1/2020.findings-emnlp.17} {Few-shot
  natural language generation for task-oriented dialog}.
\newblock In \emph{Proceedings of the 2020 Conference on Empirical Methods in
  Natural Language Processing: Findings, {EMNLP} 2020, Online Event, 16-20
  November 2020}, pages 172--182. Association for Computational Linguistics.

\bibitem[{Peng et~al.(2019)Peng, Huang, Lin, Ji, Chen, and
  Zhang}]{peng2019teacher}
Shuke Peng, Xinjing Huang, Zehao Lin, Feng Ji, Haiqing Chen, and Yin Zhang.
  2019.
\newblock Teacher-student framework enhanced multi-domain dialogue generation.
\newblock \emph{arXiv preprint arXiv:1908.07137}.

\bibitem[{Peters et~al.(2018)Peters, Neumann, Iyyer, Gardner, Clark, Lee, and
  Zettlemoyer}]{peters2018deep}
Matthew~E. Peters, Mark Neumann, Mohit Iyyer, Matt Gardner, Christopher Clark,
  Kenton Lee, and Luke Zettlemoyer. 2018.
\newblock \href {https://doi.org/10.18653/v1/n18-1202} {Deep contextualized
  word representations}.
\newblock In \emph{Proceedings of the 2018 Conference of the North American
  Chapter of the Association for Computational Linguistics: Human Language
  Technologies, {NAACL-HLT}}, pages 2227--2237.

\bibitem[{Radford et~al.(2019)Radford, Wu, Child, Luan, Amodei, and
  Sutskever}]{radford2019language}
Alec Radford, Jeffrey Wu, Rewon Child, David Luan, Dario Amodei, and Ilya
  Sutskever. 2019.
\newblock Language models are unsupervised multitask learners.
\newblock \emph{OpenAI Blog}, 1(8):9.

\bibitem[{Raffel et~al.(2020)Raffel, Shazeer, Roberts, Lee, Narang, Matena,
  Zhou, Li, and Liu}]{raffel2019exploring}
Colin Raffel, Noam Shazeer, Adam Roberts, Katherine Lee, Sharan Narang, Michael
  Matena, Yanqi Zhou, Wei Li, and Peter~J. Liu. 2020.
\newblock \href {http://jmlr.org/papers/v21/20-074.html} {Exploring the limits
  of transfer learning with a unified text-to-text transformer}.
\newblock \emph{J. Mach. Learn. Res.}, 21:140:1--140:67.

\bibitem[{Rastogi et~al.(2017)Rastogi, Hakkani-T{\"u}r, and
  Heck}]{rastogi2017scalable}
Abhinav Rastogi, Dilek Hakkani-T{\"u}r, and Larry Heck. 2017.
\newblock Scalable multi-domain dialogue state tracking.
\newblock In \emph{2017 IEEE Automatic Speech Recognition and Understanding
  Workshop (ASRU)}, pages 561--568. IEEE.

\bibitem[{Rastogi et~al.(2020)Rastogi, Zang, Sunkara, Gupta, and
  Khaitan}]{rastogi2019towards}
Abhinav Rastogi, Xiaoxue Zang, Srinivas Sunkara, Raghav Gupta, and Pranav
  Khaitan. 2020.
\newblock \href {https://aaai.org/ojs/index.php/AAAI/article/view/6394}
  {Towards scalable multi-domain conversational agents: The schema-guided
  dialogue dataset}.
\newblock In \emph{The Thirty-Fourth {AAAI} Conference on Artificial
  Intelligence, February 7-12, 2020}, pages 8689--8696. {AAAI} Press.

\bibitem[{Raux et~al.(2005)Raux, Langner, Bohus, Black, and
  Eskenazi}]{raux2005let}
Antoine Raux, Brian Langner, Dan Bohus, Alan~W Black, and Maxine Eskenazi.
  2005.
\newblock Let's go public! taking a spoken dialog system to the real world.
\newblock In \emph{Ninth European conference on speech communication and
  technology}.

\bibitem[{Roller et~al.(2020)Roller, Dinan, Goyal, Ju, Williamson, Liu, Xu,
  Ott, Shuster, Smith et~al.}]{roller2020recipes}
Stephen Roller, Emily Dinan, Naman Goyal, Da~Ju, Mary Williamson, Yinhan Liu,
  Jing Xu, Myle Ott, Kurt Shuster, Eric~M Smith, et~al. 2020.
\newblock Recipes for building an open-domain chatbot.
\newblock \emph{arXiv preprint arXiv:2004.13637}.

\bibitem[{Seneff and Polifroni(2000)}]{seneff2000dialogue}
Stephanie Seneff and Joseph Polifroni. 2000.
\newblock Dialogue management in the mercury flight reservation system.
\newblock In \emph{ANLP-NAACL 2000 Workshop: Conversational Systems}.

\bibitem[{Shannon(1948)}]{shannon48noisy}
Claude Shannon. 1948.
\newblock A mathematical theory of communication.
\newblock \emph{Bell System Technical Journal}, 27:379--423.

\bibitem[{Tanaka et~al.(2019)Tanaka, Takayama, and Arase}]{tanaka2019dialogue}
Koji Tanaka, Junya Takayama, and Yuki Arase. 2019.
\newblock Dialogue-act prediction of future responses based on conversation
  history.
\newblock In \emph{Proceedings of the 57th Annual Meeting of the Association
  for Computational Linguistics: Student Research Workshop}, pages 197--202.

\bibitem[{Tseng et~al.(2020)Tseng, Cheng, Fang, and
  Vandyke}]{tseng2020generative}
Bo{-}Hsiang Tseng, Jianpeng Cheng, Yimai Fang, and David Vandyke. 2020.
\newblock \href {https://doi.org/10.18653/v1/2020.acl-main.163} {A generative
  model for joint natural language understanding and generation}.
\newblock In \emph{Proceedings of the 58th Annual Meeting of the Association
  for Computational Linguistics, {ACL} 2020, Online, July 5-10, 2020}, pages
  1795--1807. Association for Computational Linguistics.

\bibitem[{Vaswani et~al.(2017)Vaswani, Shazeer, Parmar, Uszkoreit, Jones,
  Gomez, Kaiser, and Polosukhin}]{vaswani2017attention}
Ashish Vaswani, Noam Shazeer, Niki Parmar, Jakob Uszkoreit, Llion Jones,
  Aidan~N Gomez, {\L}ukasz Kaiser, and Illia Polosukhin. 2017.
\newblock Attention is all you need.
\newblock In \emph{Advances in neural information processing systems}, pages
  5998--6008.

\bibitem[{Wen et~al.(2017{\natexlab{a}})Wen, Miao, Blunsom, and
  Young}]{wen2017latent}
Tsung{-}Hsien Wen, Yishu Miao, Phil Blunsom, and Steve~J. Young.
  2017{\natexlab{a}}.
\newblock \href {http://arxiv.org/abs/1705.10229} {Latent intention dialogue
  models}.
\newblock \emph{CoRR}, abs/1705.10229.

\bibitem[{Wen et~al.(2017{\natexlab{b}})Wen, Vandyke, Mrksic, Gasic,
  Rojas{-}Barahona, Su, Ultes, and Young}]{wen2016network}
Tsung{-}Hsien Wen, David Vandyke, Nikola Mrksic, Milica Gasic, Lina~Maria
  Rojas{-}Barahona, Pei{-}Hao Su, Stefan Ultes, and Steve~J. Young.
  2017{\natexlab{b}}.
\newblock \href {https://doi.org/10.18653/v1/e17-1042} {A network-based
  end-to-end trainable task-oriented dialogue system}.
\newblock In \emph{Proceedings of the 15th Conference of the European Chapter
  of the Association for Computational Linguistics, {EACL} 2017, Valencia,
  Spain, April 3-7, 2017, Volume 1: Long Papers}, pages 438--449. Association
  for Computational Linguistics.

\bibitem[{Wu et~al.(2020{\natexlab{a}})Wu, Hoi, Socher, and Xiong}]{wu2020tod}
Chien{-}Sheng Wu, Steven C.~H. Hoi, Richard Socher, and Caiming Xiong.
  2020{\natexlab{a}}.
\newblock \href {https://doi.org/10.18653/v1/2020.emnlp-main.66} {{TOD-BERT:}
  pre-trained natural language understanding for task-oriented dialogue}.
\newblock In \emph{Proceedings of the 2020 Conference on Empirical Methods in
  Natural Language Processing, {EMNLP} 2020, Online, November 16-20, 2020},
  pages 917--929. Association for Computational Linguistics.

\bibitem[{Wu et~al.(2019{\natexlab{a}})Wu, Madotto, Hosseini{-}Asl, Xiong,
  Socher, and Fung}]{wu2019transferable}
Chien{-}Sheng Wu, Andrea Madotto, Ehsan Hosseini{-}Asl, Caiming Xiong, Richard
  Socher, and Pascale Fung. 2019{\natexlab{a}}.
\newblock \href {https://doi.org/10.18653/v1/p19-1078} {Transferable
  multi-domain state generator for task-oriented dialogue systems}.
\newblock In \emph{Proceedings of the 57th Conference of the Association for
  Computational Linguistics, {ACL} 2019, Florence, Italy, July 28- August 2,
  2019, Volume 1: Long Papers}, pages 808--819. Association for Computational
  Linguistics.

\bibitem[{Wu et~al.(2019{\natexlab{b}})Wu, Zhang, Li, and
  Yu}]{wu2019alternating}
Qingyang Wu, Yichi Zhang, Yu~Li, and Zhou Yu. 2019{\natexlab{b}}.
\newblock Alternating recurrent dialog model with large-scale pre-trained
  language models.
\newblock \emph{arXiv preprint arXiv:1910.03756}.

\bibitem[{Wu et~al.(2020{\natexlab{b}})Wu, Galley, Brockett, Zhang, Gao, Quirk,
  Koncel-Kedziorski, Gao, Hajishirzi, Ostendorf et~al.}]{wu2020controllable}
Zeqiu Wu, Michel Galley, Chris Brockett, Yizhe Zhang, Xiang Gao, Chris Quirk,
  Rik Koncel-Kedziorski, Jianfeng Gao, Hannaneh Hajishirzi, Mari Ostendorf,
  et~al. 2020{\natexlab{b}}.
\newblock A controllable model of grounded response generation.
\newblock \emph{arXiv preprint arXiv:2005.00613}.

\bibitem[{Yang et~al.(2021)Yang, Li, and Quan}]{yang2020ubar}
Yunyi Yang, Yunhao Li, and Xiaojun Quan. 2021.
\newblock Ubar: Towards fully end-to-end task-oriented dialog systems with
  gpt-2.
\newblock \emph{The Thirty-Fifth {AAAI} Conference on Artificial Intelligence}.

\bibitem[{Yee et~al.(2019)Yee, Dauphin, and Auli}]{yee2019simple}
Kyra Yee, Yann~N. Dauphin, and Michael Auli. 2019.
\newblock \href {https://doi.org/10.18653/v1/D19-1571} {Simple and effective
  noisy channel modeling for neural machine translation}.
\newblock In \emph{Proceedings of the 2019 Conference on Empirical Methods in
  Natural Language Processing and the 9th International Joint Conference on
  Natural Language Processing, {EMNLP-IJCNLP} 2019, Hong Kong, China, November
  3-7, 2019}, pages 5695--5700. Association for Computational Linguistics.

\bibitem[{You et~al.(2020)You, Li, Reddi, Hseu, Kumar, Bhojanapalli, Song,
  Demmel, Keutzer, and Hsieh}]{you2019large}
Yang You, Jing Li, Sashank~J. Reddi, Jonathan Hseu, Sanjiv Kumar, Srinadh
  Bhojanapalli, Xiaodan Song, James Demmel, Kurt Keutzer, and Cho{-}Jui Hsieh.
  2020.
\newblock \href {https://openreview.net/forum?id=Syx4wnEtvH} {Large batch
  optimization for deep learning: Training {BERT} in 76 minutes}.
\newblock In \emph{8th International Conference on Learning Representations,
  {ICLR} 2020, Addis Ababa, Ethiopia, April 26-30, 2020}.

\bibitem[{Yu et~al.(2017)Yu, Blunsom, Dyer, Grefenstette, and
  Kocisk{\'{y}}}]{yu2016neural}
Lei Yu, Phil Blunsom, Chris Dyer, Edward Grefenstette, and Tom{\'{a}}s
  Kocisk{\'{y}}. 2017.
\newblock \href {https://openreview.net/forum?id=SJ25-B5eg} {The neural noisy
  channel}.
\newblock In \emph{5th International Conference on Learning Representations,
  {ICLR} 2017, Toulon, France, April 24-26, 2017, Conference Track
  Proceedings}.

\bibitem[{Yu et~al.(2020)Yu, Sartran, Stokowiec, Ling, Kong, Blunsom, and
  Dyer}]{yu2020better}
Lei Yu, Laurent Sartran, Wojciech Stokowiec, Wang Ling, Lingpeng Kong, Phil
  Blunsom, and Chris Dyer. 2020.
\newblock Better document-level machine translation with bayes’ rule.
\newblock \emph{Transactions of the Association for Computational Linguistics},
  8:346--360.

\bibitem[{Zhang et~al.(2019)Zhang, Hashimoto, Wu, Wan, Yu, Socher, and
  Xiong}]{zhang2019find}
Jian-Guo Zhang, Kazuma Hashimoto, Chien-Sheng Wu, Yao Wan, Philip~S Yu, Richard
  Socher, and Caiming Xiong. 2019.
\newblock Find or classify? dual strategy for slot-value predictions on
  multi-domain dialog state tracking.
\newblock \emph{arXiv preprint arXiv:1910.03544}.

\bibitem[{Zhang et~al.(2020{\natexlab{a}})Zhang, Ou, and Yu}]{zhang2019task}
Yichi Zhang, Zhijian Ou, and Zhou Yu. 2020{\natexlab{a}}.
\newblock \href {https://aaai.org/ojs/index.php/AAAI/article/view/6507}
  {Task-oriented dialog systems that consider multiple appropriate responses
  under the same context}.
\newblock In \emph{The Thirty-Fourth {AAAI} Conference on Artificial
  Intelligence, {AAAI} 2020, The Thirty-Second Innovative Applications of
  Artificial Intelligence Conference, {IAAI} 2020, The Tenth {AAAI} Symposium
  on Educational Advances in Artificial Intelligence, {EAAI} 2020, New York,
  NY, USA, February 7-12, 2020}, pages 9604--9611. {AAAI} Press.

\bibitem[{Zhang et~al.(2020{\natexlab{b}})Zhang, Sun, Galley, Chen, Brockett,
  Gao, Gao, Liu, and Dolan}]{zhang2019dialogpt}
Yizhe Zhang, Siqi Sun, Michel Galley, Yen{-}Chun Chen, Chris Brockett, Xiang
  Gao, Jianfeng Gao, Jingjing Liu, and Bill Dolan. 2020{\natexlab{b}}.
\newblock \href {https://www.aclweb.org/anthology/2020.acl-demos.30/}
  {{DIALOGPT} : Large-scale generative pre-training for conversational response
  generation}.
\newblock In \emph{Proceedings of the 58th Annual Meeting of the Association
  for Computational Linguistics: System Demonstrations, {ACL} 2020, Online,
  July 5-10, 2020}, pages 270--278. Association for Computational Linguistics.

\bibitem[{Zhao et~al.(2019)Zhao, Xie, and Esk{\'{e}}nazi}]{zhao2019rethinking}
Tiancheng Zhao, Kaige Xie, and Maxine Esk{\'{e}}nazi. 2019.
\newblock \href {https://doi.org/10.18653/v1/n19-1123} {Rethinking action
  spaces for reinforcement learning in end-to-end dialog agents with latent
  variable models}.
\newblock In \emph{Proceedings of the 2019 Conference of the North American
  Chapter of the Association for Computational Linguistics: Human Language
  Technologies, {NAACL-HLT} 2019, Minneapolis, MN, USA, June 2-7, 2019, Volume
  1 (Long and Short Papers)}, pages 1208--1218. Association for Computational
  Linguistics.

\bibitem[{Zhou and Small(2019)}]{zhou2019multi}
Li~Zhou and Kevin Small. 2019.
\newblock Multi-domain dialogue state tracking as dynamic knowledge graph
  enhanced question answering.
\newblock \emph{arXiv preprint arXiv:1911.06192}.

\end{thebibliography}
\bibliographystyle{acl_natbib}

\end{document}